\documentclass[times, twocolumn, final]{elsarticle}

\usepackage{medima}
\usepackage{framed,multirow}
\usepackage{booktabs}       
\usepackage{amssymb}

\usepackage{url}
\usepackage{xcolor}

\usepackage{hyperref}
\usepackage{multirow}
\usepackage{bbding}
\usepackage{pifont}
\usepackage{gensymb}
\usepackage{color}
\definecolor{newcolor}{rgb}{.8,.349,.1}
\usepackage{amsmath}

\usepackage{makeidx}  
\usepackage{url}
\usepackage{amsmath}
\usepackage{amssymb}
\usepackage{latexsym}
\usepackage{subfigure}
\usepackage{graphicx}
\usepackage{multirow}
\usepackage{color}
\usepackage{hyperref}
\usepackage{marginnote}
\usepackage{lipsum}

\usepackage{enumitem}
\usepackage{bigstrut}

\definecolor{red}{rgb}{0.8,0,0}
\definecolor{blue}{rgb}{0,0,0.8}
\definecolor{green}{rgb}{0,0.4,0}

\newcommand{\change}[2]{}
\newcommand{\lchange}[2]{}

\newcommand{\changed}[3]{#3}

\usepackage{pdfpages}

\usepackage[utf8]{inputenc} 
\usepackage[T1]{fontenc}    
\usepackage{hyperref}       
\usepackage{url}            
\usepackage{booktabs}       
\usepackage{amsfonts}       
\usepackage{nicefrac}       
\usepackage{microtype}      
\usepackage{lipsum}		
\usepackage{graphicx}
\usepackage{natbib}
\usepackage{doi}

\usepackage{multirow}
\usepackage{color}
\definecolor{newcolor}{rgb}{.8,.349,.1}
\usepackage{amsmath}

\begin{document}
\journal{Medical Image Analysis}



\verso{Chuyan Zhang \textit{et~al.}}

\begin{frontmatter}

\title{Dive into the Details of Self-Supervised Learning for Medical Image Analysis}%

\author[{1}]{Chuyan \snm{Zhang}}
\author[{1}]{Hao \snm{Zheng}}
\author[{1}]{Yun \snm{Gu}}

\address[1]{Institute of Medical Robotics, Shanghai Jiao Tong University, Shanghai, China}


\begin{abstract}
Self-supervised learning (SSL) has achieved remarkable performance in various medical imaging tasks by dint of priors from massive unlabelled data. However, regarding a specific downstream task, there is still a lack of an instruction book on how to select suitable pretext tasks and implementation details throughout the standard ``pretrain-then-finetune'' workflow. In this work, we focus on exploiting the capacity of SSL in terms of four realistic and significant issues: (1) the impact of SSL on imbalanced datasets, (2) the network architecture, (3) the applicability of upstream tasks to downstream tasks and (4) the stacking effect of SSL and common policies for deep learning. We provide a large-scale, in-depth and fine-grained study through extensive experiments on predictive, contrastive, generative and multi-SSL algorithms. Based on the results, we have uncovered several insights. Positively, SSL advances class-imbalanced learning mainly by boosting the performance of the rare class, which is of interest to clinical diagnosis. Unfortunately, SSL offers marginal or even negative returns in some cases, including severely imbalanced and relatively balanced data regimes, as well as combinations with common training policies. Our intriguing findings provide practical guidelines for the usage of SSL in the medical context and highlight the need for developing universal pretext tasks to accommodate diverse application scenarios. The code of this paper can be found at~\url{https://github.com/EndoluminalSurgicalVision-IMR/Medical-SSL}.
\end{abstract}

\end{frontmatter}

\section{Introduction}\label{sec:intro}

Transfer learning from external data has become the mainstream deep learning technology to deal with the annotation scarcity in medical imaging. The de facto standard pipeline consists of two steps: 1) a Convolutional Neural Network (CNN) is first pretrained on a source dataset. 2) the pretrained network is then fine-tuned using limited annotated data with an adaptation layer added on top of its architecture for a specific target task \citep{oquab2014learning}. It is generally understood that the merits of transfer learning are contributed by the feature reuse and low-level statistics from pretrained models \citep{neyshabur2020being}. Driven by the desire to reduce manual annotations of source data, Self-Supervised Learning (SSL) is progressively overtaking Fully-Supervised Learning (FSL) to the top of the table for pretraining paradigms in state-of-the-art computer vision techniques \citep{ericsson2021well, islam2021broad,hosseinzadeh2021systematic}. 

SSL formulates pretext tasks to learn semantically useful representations from the unlabelled data, which can boost the performance of downstream tasks. The design of these pretext tasks should encourage networks to capture high-level semantics rather than trivial features~\citep{chen2019context, chowdhury2021applying}. With the impressive achievements of SSL in computer vision, there are growing appeals for developing SSL techniques tailored to medical data. Thus far, a variety of studies have demonstrated the effectiveness of SSL approaches across various medical image recognition tasks, including classification~\citep{taleb20203d, sowrirajan2021moco,zhou2021models,taher2022caid,haghighi2022dira}, object detection/localization~\citep{tajbakhsh2019surrogate, chen2019context,benvcevic2022self, haghighi2022dira}, segmentation~\citep{taleb20203d, tao2020revisiting, zhou2021models, taher2022caid, haghighi2022dira}, \changed{M2.4.1}{\link{R2.4}}{anomaly detection~\citep{tan2021detecting, Zhao2021anomaly}, few-shot learning~\citep{ouyang2020self}, image registration~\citep{li2018non}, and image denoising~\citep{xu2021deformed2self}}. 
\begin{figure*}[!ht]
    \centering
    \includegraphics[width=1.0\linewidth]{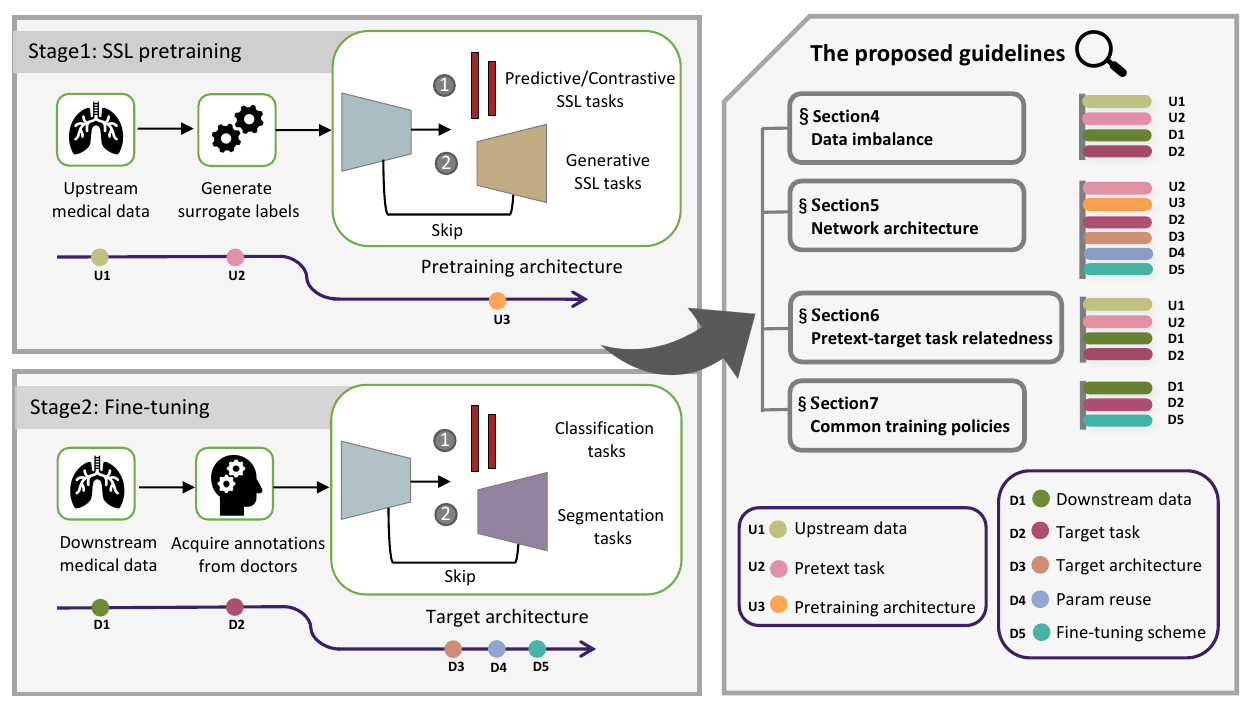}
    \caption{The overview of the general paradigm of self-supervised learning in medical imaging. It includes two training stages, namely SSL pretraining and fine-tuning. Our proposed guidelines in four dimensions (1. Data imbalance; 2. Network architecture; 3. Pretext-target task relatedness; 4. Common training policies) are remarked throughout the training process, involving data, models, tasks and transfer schemes. Specifically, the data imbalance and pretext-target task relatedness involve the upstream and downstream data and tasks. The network architecture is related to the pretraining and target networks, tasks, parameter reuse as well as the fine-tuning scheme. The common training policies for solving target tasks are associated with the downstream data, target task and fine-tuning scheme.}
    \label{fig:method_overview}
\end{figure*}

In the context of medical imaging, SSL literature mainly focuses on creating novel pretext tasks and validating the improved performance of the self-supervised learned features. Recently, several systematic reviews have been undertaken for SSL \citep{xu2021review,chowdhury2021applying, shurrab2022self,chen2022recent}. In addition to the performance evaluation, a few works discovered broader implications of SSL approaches, such as better robustness and generalization \citep{navarro2021evaluating, srinivasan2021robustness}. Despite these preliminary attempts, there are still underexplored problems in the workflow of ``pretrain-then-finetune". Firstly, given diverse pretext tasks with substantially different formulations, e.g. generative learning and contrastive learning \citep{liu2021self}, which type of method is empirically effective? This is hard to assess from current literature with potential experimental flaws, including the incomplete comparisons of limited types of SSL methods, inconsistent validity reported by the same method in different evaluation conditions and the biased selection of baselines. Secondly, data bias is a prominent issue in medical images since malignant samples are always in the minority. In regular CNN training, data resampling and augmentations are widely adopted to improve class-imbalanced learning from the perspective of expanding the frequency of the rare class. However, little attention has been paid to the behaviours of SSL algorithms under data imbalance. Furthermore, to get a clear picture of the utilities that SSL can provide, there is a necessity to investigate the additive effect of SSL with data resampling or augmentations. Thirdly, the success of SSL-based transfer learning depends on two imperative implementations: how to design a proper pretext task, and how to transfer useful representations from the pretrained model to downstream tasks? For the former question, inappropriate implementations of the pretext task might cause the network finds trivial solutions as a shortcut to accomplish its goals. For the latter question, it is essential to determine whether the self-supervised learned features are preserved active in downstream learning or not.

In this paper, we precisely fill this research gap by taking a closer look at the aforementioned problems. We first revisit prevalent SSL techniques covering adequate types and introduce an evaluation benchmark across three popular medical datasets for the underexplored issues in the existing literature. Based on extensive experiments (nearly 250 experiments, 2000 GPU hours), we perform a fine-grained study and propose a series of guidelines for the community in order to exploit the capacity of SSL. Fig.~\ref{fig:method_overview} navigates the studied topics throughout pretraining and fine-tuning, which can actually serve as a guidebook. As a result of this study, we uncover several crucial insights. The most significant observations are summarized as follows:

\begin{itemize}
	\item SSL facilitates class-imbalanced problems with remarkable improvement in the minority class but marginal gains or occasional losses in the majority class. 
	\item The data bias in pretraining data also affects the representation learning, leading to poor target performance. Using data resampling in the fine-tuning can neutralize such skewed representations and yield mutual benefits with SSL pretraining.
	\item For encoder-decoder architectures, the representations in the pretrained encoder are more meaningful than the decoder as the decoder might impose the risk of overfitting inductive information for solving the pretext task itself.
	\item SSL pretraining only offers substantial gains in the absence of data augmentation for training target tasks. When using strong data augmentation, most of the SSL methods hurt the segmentation performance, which highlights the need to rethink the value of SSL pretraining in medical segmentation tasks.
\end{itemize}

\changed{M1.3}{\link{R1.3}}{The remainder of this paper is organized as follows: Section \ref{sec:review} briefly reviews previous SSL literature in medical image analysis. The overall study outline, main formulations of the selected SSL tasks, experiment configurations and preliminary analysis are introduced in Section \ref{sec:method}. Then, we explore the behaviours of diverse SSL methods in terms of four aspects: data imbalance, network components, pretext-target task relatedness and common training policies, sequentially in Section \ref{sec:imbalance}, Section \ref{sec:u-shape}, Section \ref{sec:pretexttasks} and Section \ref{sec:augmentation}. In Section \ref{sec:discussion}, we conduct detailed discussions to answer the questions raised in Section \ref{sec:intro} and suggest promising future directions for SSL in medical imaging. The conclusions are finally presented in Section \ref{sec:conclusion}.}

\section{Self-Supervision Literature Review}
\label{sec:review}

To relieve the issue of scarce annotations, SSL works are increasingly proposed for medical image analysis. This review is organized into three parts: firstly, we present a formal concept of SSL and review previous SSL approaches in terms of four categories during the years 2019 to 2022. Secondly, we summarize previous literature reviews. Thirdly, we also survey some analytical works on deeply understanding SSL. In the final part, we distinguish our work from recent relevant papers. \changed{M1.2}{\link{R1.2}}{Table~\ref{tab:analytical} summarizes SSL studies to be discussed in this section in terms of research topic, research form as well as image domain.}
 

\subsection{Self-supervised approaches in medical image analysis}\label{sec:ours_review}

\textbf{Formal concept of SSL.} The core idea of SSL is to learn useful representations from the intrinsic information in unlabelled images by solving a pretext task (also called the auxiliary or proxy task). According to the purpose of a certain pretext task, surrogate supervisory signals are automatically generated from an unlabelled dataset which is called the source dataset or upstream dataset. Instead of the pretext task itself, the real research interest is in downstream tasks, e.g. image classification or semantic segmentation. In the notion of transfer learning, SSL pretraining is adopted as an alternative scheme to improve the performance of downstream tasks. When transferred to a specific downstream task, the SSL pretrained model is fine-tuned with a labelled dataset, called the target dataset or downstream dataset. According to the patterns of pretext tasks, SSL methods can be categorized into predictive SSL, generative SSL, contrastive SSL and multi-SSL.
\begin{table*}[!t]
\renewcommand{\arraystretch}{0.7}
\centering
\caption{The overview of SSL literature discussed in Section \ref{sec:review}. Here Emp. denotes empirical study and Theor. denotes theoretical study.}
 \resizebox{0.92\textwidth}{!}{
\begin{tabular}{ccccc}
\toprule
SSL field                    & Research form                                                 & Image domain             & Research topic                                                                                                                       & Literature                                                                                                                                                                                                                                                      \\ \midrule &  & &  & \cite{ li2021novel, zhu2020rubik, bai2019self, tan2021detecting} \\
& & & Predictive SSL&\cite{blendowski2019learn, zhuang2019self, taleb20203d} \\
& & & &\cite{nguyen2020self, tajbakhsh2019surrogate, ouyang2020self}\\ 
 \cmidrule{4-5} 
                      Methodology       &    Emp.                                                                  &   Medical                         & Generative SSL                                                                                                                      & \begin{tabular}[c]{@{}c@{}}\cite{tajbakhsh2019surrogate,chen2019context, zhou2021models}\\  \cite{tao2020revisiting,Zhao2021anomaly, xu2021deformed2self}\end{tabular}               
               \\ \cmidrule{4-5} 
                             &                                                                     &                          & Contrastive SSL                                                                                                                      & \begin{tabular}[c]{@{}c@{}}\cite{sowrirajan2021moco, sriram2021covid,vu2021medaug}\\  \cite{azizi2021big, chaitanya2020contrastive, zhou2020comparing}\end{tabular}  \\\cmidrule{4-5}
                             &                                                                     &                          & Multi-SSL                                                                                                                      & \begin{tabular}[c]{@{}c@{}}\cite{haghighi2020learning, zhang2021sar,dong2021self}\\  \cite{zhou2021preservational, taher2022caid, haghighi2022dira}\end{tabular}
                             \\\midrule
Review                       & —                                                               & Medical                  & —                                                                                                                    & \begin{tabular}[c]{@{}c@{}}\cite{xu2021review, chowdhury2021applying}\\ \cite{shurrab2022self, chen2022recent}\end{tabular}                                                                                                                                       \\ \midrule

    & Emp.                                                           & Medical                  & Robustness                                                                                                                           & \cite{srinivasan2021robustness, navarro2021evaluating}  \\ \cmidrule{2-5} 
& Emp.\& Theor. & Non-medical& Data imbalance&\cite{liu2021imbalance}\\ \cmidrule{2-5} Analysis & & & Data imbalance&\\
&Emp.\& Theor. &Medical &Network architecture &Ours\\
& & & Pretext-target task relatedness &\\
& & & Common training policies &\\\bottomrule
\end{tabular}}
\label{tab:analytical}
\end{table*}

\textbf{Predictive SSL: predicting transformations or pseudo labels.} Most of the predictive tasks learn the contextual or structural semantics via predicting spatial transformations of images. In these methods, the pretext task is formulated as a classification problem, mapping inputs to a scalar or vector in the semantic space. The common transformation-based predictive tasks consider the relative positions~\citep{doersch2015unsupervised}, the rotation angles \citep{komodakis2018unsupervised} and the jigsaw puzzles \citep{noroozi2016unsupervised} etc. A majority of works have proved the effectiveness of these traditional pretext tasks in medical imaging: 
\cite{tajbakhsh2019surrogate} demonstrated that utilizing rotation prediction as a pretraining scheme for medical imaging tasks is more effective than training from scratch, particularly under low-annotation data regimes. \cite{taleb20203d} emphasized the superiority of 3D medical models over 2D models and presented 3D versions of relative patch location, rotation prediction and jigsaw puzzles. Other works developed novel predictive SSL tasks for medical scenarios: \cite{bai2019self} proposed to predict anatomical positions for cardiac magnetic resonance (MR) image segmentation. \cite{blendowski2019learn} presented a new SSL approach by predicting orthogonal patch offsets. \cite{zhuang2019self} designed a novel proxy task named ``Rubik's cube recovery'': partitioning a 3D volume into cubes in a grid, followed by rearranging and randomly rotating cubes, and then predicting the cube orders and rotations in a unified framework. \cite{zhu2020rubik} further extended ``Rubik's cube recovery'' by adding a new sub-task of identifying whether a cube is masked or not. \cite{nguyen2020self} exploited the spatial information inherent in medical data by corrupting patches from a small range of slices and training a model to predict whether the slice is corrupted or not as well as the slice index. \cite{li2021novel} introduced two alternative auxiliary tasks into the multiple instance learning framework for COVID-19 diagnosis: the relative patch location prediction  of two patches from the same slice, and the absolute patch location prediction of a single patch. Another line of work learns more fine-grained representations through pixel-level pseudo labels. \changed{M2.4.2}{
\link{R2.4}}{\cite{ouyang2020self} generated pseudo labels on super-pixels for few-shot medical image segmentation. \cite{tan2021detecting} synthesized defects by Poisson interpolation and drove the network to detect these anomalies.}

\textbf{Generative SSL: recovering the corrupted or masked part of the input.} The underlying idea of generative pretext tasks is to learn context-aware features by recovering the distorted or unobserved image crops. Generative SSL demands an encoder to extract compressed representations from input pixels and a decoder to convert the latent representations back to pixels. The reconstruction objective is always the original data while diversified transformations are applied to the original data as the input, such as inpainting \citep {pathak2016context} and colorization \citep {zhang2016colorful, larsson2017colorization}. Inspired by prevalent generative tasks in computer vision, \cite{tajbakhsh2019surrogate} implemented colorization in a conditional Generative Adversarial Network (GAN) \citep{larsson2017colorization} and patch reconstruction in a Wasserstein GAN \citep{arjovsky2017wasserstein} for medical imaging tasks. \cite{chen2019context} claimed that existing pretext tasks such as local context prediction \citep{pathak2016context} often yield marginal improvements on medical datasets. To better exploit the rich information in unlabelled images, they proposed a new SSL strategy for context restoration, in which small patches are randomly switched in an image. As an extension of previous context-restoration-based works, \cite{zhou2021models} integrated four data transformations (non-linear transformation, local-shuffling, outer-cutout, and inner-cutout) into a unified reconstruction model, namely ``Model Genesis''. Based on Rubik's cube \citep{zhuang2019self, zhu2020rubik}, \cite{tao2020revisiting} developed a new generative task of recovering the original image from the disrupted version and an adversarial task of distinguishing whether the cubes are in the correct arrangement. 
Specifically, the individual change of a cube is forbidden and the movement of cubes is bound to its adjacent cubes.  \changed{M2.4.3}{
\link{R2.4}}{\cite{Zhao2021anomaly} present a reconstruction-based SSL technique with a structure similarity loss and a center constraint loss in a GAN framework for anomaly detection of medical images. \cite{xu2021deformed2self} proposed a self-supervised denoising framework where a single-image denoising network is first trained to recover missing pixels from randomly masked input images and then a multi-image denoising network is built to improve image quality by exploiting the similarity of the image content among different time frames.}

\textbf{Contrastive SSL: learning maximum agreement between two views of one sample in the embedding space.} Contrastive learning aims at maximizing the mutual information between positive image pairs and minimizing the representation similarity of negative image pairs if necessary. In general, the positive pairs are two augmented views of one instance while the negative pairs are from different instances. The network can therefore capture discriminative representations of instances that are valuable for pattern recognition. 
Recently, contrastive learning has become one of the most popular trends in SSL and has shown outstanding performance in a wide range of computer vision tasks, leading to a proliferation of related studies \citep{chen2020simple, he2020momentum, chen2020improved, grill2020bootstrap, chen2021exploring}. 
\changed{M2.5.1}{\link{R2.5}}{In contrastive learning, the knowledge embedded in the learned representations is dominated by the selection of positive and negative pairs. Thus, the pair generation proposed for object-centric natural images in mainstream contrastive learning methods, involving data augmentations and pair mining strategies, could lead to meaningless representations for medical images with complex semantic concepts: (1) There is always a diverse set of object-of-interest and much more clinically-related background information in a medical image. (2) The consistent anatomical structures in medical images cause subtle instance-wise differences, e.g. Magnetic Resonance Imaging (MRI) and Computed Tomography (CT) scans. To address such issues, many efforts have been made to carefully design pair selection strategies based on prevalent contrastive learning frameworks to preserve the pathological semantics in the images, thereby significantly improving the performance of medical datasets over the classic counterparts.}
These works can be further divided into two groups: global-attention contrastive SSL and local-attention contrastive SSL. For global-attention methods, \cite{sowrirajan2021moco} claimed that the original augmentations in MOCO \citep{he2020momentum} are inappropriate for gray-scale medical images. For instance, random crop and blurring might eliminate the lesions. Therefore, they proposed the MoCo-CXR to fit into chest X-Ray images by modifying augmentations in MoCo. As another extension work of MoCo, \cite{sriram2021covid} performed SSL pretraining on non-COVID datasets to benefit downstream COVID-19 predictions. Based on MoCo-CXR, \cite{vu2021medaug} put forward an SSL method named MedAug which generates positive pairs from different images of one patient according to the metadata. One similar work to MedAug was proposed by \cite{azizi2021big}, where they utilized SimCLR \citep{chen2020simple} as the benchmarking framework and introduced a Multi-Instance Contrastive Learning (MICLe) to construct more informative positive pairs from multiple images of the same patient. For local-attention methods,  \cite{chaitanya2020contrastive} made two improvements to SimCLR for 3D medical image segmentation. They devised a novel contrasting strategy to harness the structural similarity of volumetric medical images and proposed a local contrast loss to learn more fine-grained representations. \cite{zhou2020comparing} mixed images and features to generate homogeneous and heterogeneous pairs and built a Comparing to Learn (C2L) framework for radiograph tasks. 

\textbf{Multi-SSL: combining multiple SSL pretext tasks into one framework.} Due to the limitation that a single pretext task could learn task-biased features, multi-SSL adopts different self-supervision signals for network training to extract robust representations. In this section, we only review the SSL works that combine pretext tasks of different types (predictive, generative and contrastive). Based on Model Genesis, \cite{haghighi2020learning} proposed a Semantic Genesis framework to learn semantics-enriched representation by three modules: self-restoration, self-discovery and self-classification. 
In the self-discovery module, semantically similar samples are discovered by comparing the representations learned by an autoencoder. Then, they assign anatomical pattern labels to patches at fixed positions. The self-classification and self-recovery refer to the pattern classification task and the reconstruction task used in Model Genesis, respectively. As another extension work of Model Genesis, \cite{zhang2021sar} added a scale-aware proxy task of predicting the input's scale into the Model Genesis framework to learn multi-level representations. \cite{zhou2021preservational} incorporated contrastive and generative SSL into a PCRL framework, where "preservational learning" is proposed for the generative SSL to preserve more information. \cite{dong2021self} designed a multi-task framework consisting of contrastive and generative sub-tasks to collaboratively learn instance discrimination and the underlying structures from sequential 2D medical data. 
Similarly, \cite{taher2022caid} argued that state-of-the-art contrastive SSL algorithms learns global instance discrimination of photographic images while medical image analysis requires more attention to local features. Inspired by PCRL, they added a generative task into the contrastive learning framework to obtain more fine-grained features. To benefit from collaborative learning, \cite{haghighi2022dira} developed DiRA by integrating discriminative, restorative, and adversarial learning.

\subsection{Previous reviews of self-supervised learning}
Thus far, a series of studies have reviewed existing SSL approaches for medical images \citep{xu2021review,chowdhury2021applying, shurrab2022self,chen2022recent}. Most of the reviews were organized into an analogical structure consisting of the introduction, survey and discussion parts. The introduction part always includes the background of SSL and the selection criteria of published papers to be reviewed. The survey part summarizes numerous related works. In the end, the latest research trends, potential issues and future directions will be fully discussed to offer fresh perspectives on the community.

\subsection{Analytical work for understanding self-supervised learning}
Despite the success of SSL in improving the performance of downstream tasks, the reasons behind the benefits of SSL remain largely underexamined. To address this need, there is a growing body of literature analyzing the properties of SSL in different settings through carefully designed experiments. 
\cite{navarro2021evaluating} pointed out that self-supervision only yields marginal performance improvements in medical datasets compared with full-supervised learning and they thus turned to assess the robustness and generalization of SSL algorithms. The evaluations on disrupted images finally revealed the advantages of SSL in acquiring robust representations. \cite{srinivasan2021robustness}
paid much attention to the broader implications of SSL in the context of diabetic retinopathy, involving the specificity, interpretability and robustness of the self-supervised learned features. \cite{liu2021imbalance} explored the behaviours of SSL in data-imbalanced scenarios and discovered that the self-supervised learned representations are more robust to class imbalance than supervised learned representations.

\subsection{Position of our work}
Currently, a few excellent reviews have established a comprehensive body of knowledge on SSL. Furthermore, some analysis approaches help researchers to understand the role of SSL from certain perspectives, e.g. robustness and generalization. However, choosing a conducive pretext task and executing it appropriately on a target task remains a complex problem for researchers.
In this work, we briefly review relevant literature but concentrate more on the analysis part. 
Different from previous analytical works towards investigating one predominant property of SSL with limited methods (sometimes encompassing only a single type of SSL like the contrastive SSL), we aim to consider a set of possible issues in practice and present detailed references on sufficient types of SSL for researchers. Based on several crucial observations in extensive experiments, we discuss current limitations and promising future directions of SSL in the field of medical imaging.

\section{Methodology and comprehensive comparison}
\label{sec:method} 
In this study, a brief outline is first provided to uncover the significant topics to be explored throughout SSL applications. Then, we illustrate the selection of SSL methods. Afterwards, the experimental setup on relevant medical datasets is described. Then, we point out the potential flaws in the experiments reported in previous works. By implementing a unified benchmarking framework, a fair comparison of representative SSL algorithms is performed across multiple medical tasks.

\subsection{Study Outline}
The study is conducted with the issues listed in Table~\ref{tab:overview_of_exp}, including four major aspects involved in the upstream and downstream training phases. Each aspect is carefully examined and analyzed with selected SSL methods on multiple datasets. We include appropriate medical tasks for exploring various factors. For instance, the class-imbalance ratio in classification tasks is easier to adjust than in segmentation tasks. Thus for simplicity, we only perform the relevant experiments in the classification task. Based on extensive experiments, we finally provide detailed guidelines for a deeper understanding and more effective usage of SSL pretraining. 

\begin{table*}[!t]
	\renewcommand{\arraystretch}{0.7}
	\centering
	\caption{The outline of our study. We conduct a fine-grained study on the impact of SSL in four aspects, covering a variety of topics. For each topic, we display the target task/tasks, which is/are selected considering the feasibility and simplicity.}
\resizebox{0.75\textwidth}{!}{
	\begin{tabular}{cccc}
	\toprule
 & 
	  \multirow{2}{*}{Research object}                                                                         & \multicolumn{2}{c}{Target task}                  \\ \cmidrule(r){3-4} 
  &

  & Classification & Segmentation \\ \toprule
	\multicolumn{1}{c}{}                                                & Degrees of data imbalance                                  & \multicolumn{1}{c}{\checkmark}              & \XSolidBrush            \\ 
	\multicolumn{1}{c}{\multirow{-2}{*}{\textbf{Data imbalance}}}       & \multicolumn{1}{l}{Performance of different classes} & \multicolumn{1}{c}{\checkmark}              & \checkmark            \\ \midrule 
	\multicolumn{1}{c}{}                                                & Encoder                                               & \multicolumn{1}{c}{\checkmark}              & \checkmark             \\ 
	\multicolumn{1}{c}{}                                                & Decoder                                               & \multicolumn{1}{c}{\XSolidBrush}              & \checkmark             \\ 
	\multicolumn{1}{c}{}                                                & Skip connections                                      & \multicolumn{1}{c}{\checkmark}              & \checkmark              \\ 
	\multicolumn{1}{c}{}                                                & Network capability                                             & \multicolumn{1}{c}{\XSolidBrush}              & \checkmark     
 \\  
	 \multicolumn{1}{c}{\multirow{-3}{*}{\textbf{Network architecture}}} &
  
 Normalization layers                                  & \multicolumn{1}{c}{\checkmark}              & \checkmark  \\ \multicolumn{1}{c}{\multirow{-1}{*}} &Fine-tuning strategy                                  & \multicolumn{1}{c}{\checkmark}              & \checkmark            \\ \midrule 
	\multicolumn{1}{c}{}                                                & Selection of pretext tasks                            & \multicolumn{1}{c}{\checkmark}              & \checkmark             \\ 
	\multicolumn{1}{c}{}                                                & Input data size                                       & \multicolumn{1}{c}{\checkmark}              & \checkmark            \\ 
	\multicolumn{1}{c}{\multirow{-3}{*}{\textbf{Pretext-target task relatedness}}}        & Feature  analysis                            & \multicolumn{1}{c}{\checkmark}              & \multicolumn{1}{c}{\checkmark}      \\ \midrule       
	\multicolumn{1}{c}{\multirow{2}{*}{\textbf{Common training policies}}}                      & Additive effect with data resampling               & \checkmark       & \multicolumn{1}{c}{\XSolidBrush}   \\ \multicolumn{1}{c}{} & Additive effect with data augmentation                          & \multicolumn{1}{c}{\XSolidBrush}              & \checkmark           \\
 \bottomrule
\end{tabular}}
\label{tab:overview_of_exp}
\end{table*}

\begin{table*}[!t]
	\renewcommand{\arraystretch}{0.7}
	\setlength\tabcolsep{1.6pt}
\caption{Statistics of CT datasets in existing SSL works. Here Cls. denotes classification, Seg. denotes segmentation and Loc. denotes localization. We rank the datasets according to the usage frequency in publications.}
\centering
\resizebox{0.75\textwidth}{!}{
\begin{tabular}{ccccc}
	\toprule
	Name                       & \# of annotated scans & Object                    & Task              & Publication    \\ \toprule
	\multirow{4}{*}{LiTS 2017 \citep{LiTS}} & \multirow{4}{*}{130}       & \multirow{4}{*}{Liver}  & \multirow{4}{*}{Seg.} &  \cite{zhou2021models}             \\
								&                              &                       &                       &  \cite{haghighi2020learning}              \\
								&                              &                       &                       &  \cite{zhou2021preservational}            \\
								&                              &                       &                       & \cite{xie2020pgl}            \\ \midrule
	\multirow{4}{*}{LUNA 2016 \citep{LUNA16}} & \multirow{4}{*}{888} & \multirow{4}{*}{Lung nodule}  & \multirow{4}{*}{Cls.} &  \cite{zhou2021models}                         \\
								&                              &                       &                       & \cite{haghighi2020learning}              \\
								&                              &                       &                       & \cite{zhou2021preservational}         \\  
								&                              &                       &                       & \cite{tajbakhsh2019surrogate}\\\midrule
	\multirow{2}{*}{LIDC-IDRI \citep{LIDC}} & \multirow{2}{*}{1018} & \multirow{2}{*}{Lung nodule} & \multirow{2}{*}{Seg.} & \cite{zhou2021models}                         \\
								&                              &                       &                       & \cite{haghighi2020learning}              \\\midrule
								LIDC-IDRI \citep{LIDC}              & 355                    & Lung lobe                  & Seg.                  & \cite{tajbakhsh2019surrogate}       \\ \midrule
	MSD-Pancreas \citep{MSD}              & 420                     & Pancreas                   & Seg.                  & \cite{taleb20203d}        \\ \midrule
	Abdominal multi-organ \citep{Multi-organ}     & 150               & Abdominal organs                 & Loc.                  & \cite{chen2019context}\\ \midrule
	NIH-Pancreas \cite{NIH-Pancreas}             & 82                      & Pancreas                 & Seg.                  & \cite{tao2020revisiting}         \\ \midrule
	PE-CAD  \citep{PE-CAD}                   & 121             & Pulmonary emboli                  & Cls.                  & \cite{zhou2021models}            \\ \bottomrule
	\end{tabular}
 }
	\label{tab:datasets}
	\end{table*}


\subsection{Selected SSL Methodology}
In this paper, we select ten prevalent SSL approaches, adequately including the state-of-the-art methods for medical applications and spanning all the SSL types we defined before. In total, five of these ten methods are predictive SSL (ROT~\citep{komodakis2018unsupervised}, RPL~\citep{doersch2015unsupervised}, Jigsaw~\citep{taleb20203d}, RKB~\citep{zhuang2019self} and RKB+~\citep{zhu2020rubik}),
 three are generative SSL (AE~\citep{SCHMIDHUBER201585}, MG~\citep{zhou2021models}, PCRL~\cite{zhou2021preservational}\footnote{PCRL is categorized into Multi-SSL above.}) and two are contrastive SSL (SimCLR~\citep{chen2020simple}, BYOL~\citep{grill2020bootstrap}). Each of these is formally presented below.

\textbf{Rotation Prediction (ROT)} 
First proposed by \cite{komodakis2018unsupervised} for 2D images, this pretext task learns representative features by predicting rotation angles from rotated inputs. The underlying idea is that for estimating orientations, the network has to capture recurrent structures, which is meaningful for object recognition. To harness the full spatial context of 3D medical volumes, \cite{taleb20203d} modified the original task to a 3D version and framed the task as a 10-way classification problem: the input images are randomly rotated by one of $\{0\degree, 90\degree, 180\degree, 270\degree\}$ along an arbitrary axis in $\{x, y, z\}$, leading to 10 possible rotation strategies. We assume that $K$ is the number of categories, the one-hot label is $y$ and the network prediction is $p$. The objective function is a simple cross-entropy loss:
\begin{equation}       
L_{CE}=-\sum\nolimits_{j=1}^Ky_{j} \log{p_{j}} 
\label{eq:cross-entropy}
\end{equation}


\textbf{Relative Patch Location (RPL)}
Originally proposed in \cite{doersch2015unsupervised}, this pretext task seeks to learn semantic representations by predicting the spatial relations of patches in one image. In the 3D counterpart presented by \cite{taleb20203d}, the input volumetric data is first partitioned into a grid of $N$ cubes ${x_i}, i= 1...N$ and then the central cube is regarded as a reference patch $x_c$ while a query patch $x_q$ is randomly chosen from the remaining $N-1$ patches. Obviously, there are $N-1$ location relations between $x_c$ and $x_q$. Therefore, this task can be considered as a $N-1$-way classification problem. The loss function is the cross-entropy loss as written in Eq.~\ref{eq:cross-entropy}.

\textbf{Jigsaw puzzle solving (Jigsaw)}
Following the intuition behind RPL, \cite{noroozi2016unsupervised} proposed Jigsaw puzzle solving as a more complex SSL task for 2D images. The input image is first split into a grid of $n\times{n}$ patches and then permuted in random order. The goal of this task is to predict the right arrangement of shuffled patches. For a 3D image, there exist $n\times{n}\times{n}$ patches and $n^3!$ possible arrangements that are oversize for network predictions. Therefore, \cite{taleb20203d} proposed a simplified scheme of selecting $K$ possible orders with the largest Hamming distance, thereby casting the task to a K-way classification problem. The loss function is the cross-entropy loss as written in Eq.~\ref{eq:cross-entropy}. 

\textbf{Rubik's Cube (RKB)}
\cite{zhuang2019self} developed a novel 3D-based pretext task for volumetric medical data, called Rubik's cube recovery. To learn both translational and rotational invariant representations, they combined two sub-tasks, namely cube rearrangement prediction and cube rotation prediction, into a joint learning framework. For an input volume, $n\times{n}\times{n}$ shuffled cubes are generated in the same way as Jigsaw. Then, each cube $x_i, i=1... N$ is randomly rotated by $180\degree$ vertically and horizontally. Hence, there are three supervision signals: a $K$-dim one-hot label $y^{order}$ for cube orders, a $N$-dim multi-hot label $y^{hor}$ for vertical rotation, with 1 on the positions of vertically rotated cubes and 0 vice versa, and a $N$-dim multi-hot label $y^{ver}$ for horizontal rotation. 
The loss function $L_{order}$ for the cube ordering sub-task is the same as Eq.~\ref{eq:cross-entropy}. Each of the two cube rotation sub-tasks can be solved as a multi-label binary classification problem. Let $p^{hor}$ and $p^{ver}$ denote the horizontal and vertical rotation prediction, the loss function $L_{rot}$ is:
\begin{equation}       
	L_{rot}=-\sum\nolimits_{i=1}^N(y_{i}^{hor}\log{p_{i}^{hor}} + y_{i}^{ver}\log{p_{i}^{ver}})
	\label{eq:L_rot}
	\end{equation}
 
In total, the loss function comprises two terms with $\alpha$ and $\beta$ as the weights:
\begin{equation}       
L_{RKB}= {\alpha}L_{order} + {\beta}L_{rot}
\end{equation}

\textbf{Rubik's Cube+ (RKB+)}
Following the idea that a harder pretext task often enables more robust feature representations \citep{wei2019iterative}, \cite{zhu2020rubik} added a new sub-task of cube masking identification to RKB. After the cube transformations in RKB, each cube is then randomly masked with a possibility of 0.5 at each voxel. In addition to the two sub-tasks in RKB, the network is trained to identify whether each cube is masked or not, which can be cast as a multi-label binary classification problem. The ground truth for cube masking is a $N$-dim multi-hot label $y^{mask}$, with 1 on the positions of masked cubes and 0 vice versa. Let $p^{mask}$ denote the network prediction, the loss function is defined as:
\begin{equation}       
	L_{mask}=-\sum\nolimits_{i=1}^Ny_{i}^{mask}\log{p_{i}^{mask}}
	\end{equation}
 
Finally, the total loss function for joint training is composed of three parts with $\alpha$, $\beta$ and $\gamma$ as the weights:
\begin{equation}       
	L_{RKB+}= {\alpha}L_{order} + {\beta}L_{rot} + {\gamma}L_{mask}
	\end{equation}
where the formulations of $L_{order}$ and $L_{rot}$ can be found in Equation~\ref{eq:cross-entropy} and \ref{eq:L_rot}.

\textbf{Auto Encoder (AE)}
Auto Encoder \citep{SCHMIDHUBER201585} is a classical neural network architecture for learning compressed features from unlabelled data. The standard framework consists of an encoder $E$ for condensing information and a decoder $D$ for reconstructing the original input. 
We take $x$ as the input, then the loss function is expressed as:
\begin{equation}       
L_{AE}= \Vert x-D(E(x)) \Vert_2^2
\end{equation}

\textbf{Model Genesis (MG)}
Based on AE, \cite{zhou2021models} proposed an SSL framework called ``Model Genesis''. In Model Genesis, four strong transformations are randomly applied to the original medical image to generate the input, including the non-linear transformation, local-shuffling, outer-cutout and inner-cutout. Let $\mathcal{T}$ denote the transformations for $x$, the objective function is written as:
\begin{equation}       
	L_{MG}= \Vert x-D(E(\mathcal{T}(x)))  \Vert_2^2
	\end{equation}

\textbf{PCRL}
PCRL \citep{zhou2021preservational} is a combination of contrastive and generative SSL. One important innovation is that the generative pretext task recovers a transformed input with a given indicator vector, which can encourage the network to encode richer information. Moreover, a mix-up strategy is adopted to restore diverse images. Due to the great complexity of this composite task, we would not list optimization functions here. 

\textbf{SimCLR}
The mining scheme of positive and negative pairs is a crucial component in contrastive learning approaches. In SimCLR \citep{chen2020simple}, given a batch of samples ${x_i}, i=1...N$, two augmented views of one image are regarded as the positive pair $(x{_i}, x{_i}^+)$ while other different images $(x_i, x{_j}^-), i{\neq}j, i, j=1...N$ are exploited as negative pairs. The common InfoNCE loss \citep{oord2018representation} for contrastive learning is defined as:
\begin{equation}       
	L_{SimCLR}= -log\frac{exp(g(x){\cdot}g(x^+))}{exp(g(x){\cdot}g(x^+))+\sum\nolimits_{x^-}exp(g(x){\cdot}g(x^-))}
	\end{equation}
in which $g(\cdot)$ represents the corresponding embedding.

\textbf{BYOL}
Different from traditional contrastive learning methods that require sufficient negative samples, BYOL proposed by \cite{grill2020bootstrap} discards negative sampling and only relies on positive-pair learning. The architecture is composed of an online network and a target network, where the former is updated normally by gradient descent and the latter is updated by the exponential moving average of the online network parameters. Two augmented views $(x, x^+)$ of one image are passed to online and target networks respectively to obtain the embeddings $(g_o(x), g_t(x^+))$. For a positive pair, the online network is trained to predict $g_o(x)$ close to the embedding from the target network $g_t(x^+)$:
\begin{equation}       
	L_{BYOL}= \Vert g_o(x)-g_t(x^+) \Vert_2^2
\end{equation}

\subsection{Medical datasets and tasks}
We consider both 2D and 3D medical imaging tasks for experiments. First, we collect 3D CT datasets that were used as downstream tasks in published SSL works in Table~\ref{tab:datasets}. According to the rankings of publication numbers, we choose the top three datasets for 3D target tasks: LiTS, LUNA 2016 and LIDC-IDRI. Note that the training data in MSD-Liver \citep{MSD} is from LiTS. For simplicity, we use LiTS for liver segmentation and MSD for liver-tumour segmentation, though they are composed of the same CT volumes. Following \cite{zhou2021models}, we adopt 623 Chest CT scans in LUNA 2016 for 3D pretraining as well and configure the same training and test protocols for the selected 3D downstream datasets, giving rise to three target tasks: LCS (Liver segmentation in LiTS dataset), NCS (Lung nodule segmentation in LIDC-IDRI dataset) and NCC (Lung nodule false positive reduction in LUNA 2016 dataset). 

\changed{M2.1}{\link{R2.1}}{In 2D tasks, we make experiments on two retinal fundus datasets: EyePACS~\citep{eyepacs} and DRIVE~\citep{drive}. Based on the image quality labels in \cite{voets2019reproduction}, 28k gradable fundus images are chosen for 2D pretraining. The 2D target tasks include EPC (diabetic retinopathy grade classification in EyePACS dataset) and DVS (retinal vessel segmentation in DRIVE dataset). In EPC, 10\% of the entire EyePACS dataset (3511/1090/4265) is used for training/validation/testing by random sampling. In DVS, 20/5/15 retinal images are used for training/validation/testing.}

To avoid the test-image leakage between proxy and target tasks, we keep non-overlap between the pretraining data and the test data of target tasks.

\subsection{Fair Comparisons and Reproducibility}\label{sec:fair}

In the field of SSL, many works validate the proposed method by comparing it with other approaches in their experimental setup, where training from scratch is regarded as the baseline. Yet, to our knowledge, the comparison results can be biased due to the study settings.

One particular phenomenon is that an SSL approach might show a significant performance improvement in the setting of a paper but a marginal gain in the setting of another paper even on the same source and target datasets. For example, in a setting with LUNA 2016 as the source dataset and LiTS as the target dataset, MG outperforms 5\% $\sim$ 7\% than baseline in Intersection over Union (IoU) score in \cite{zhou2021models} while only 0.6\% than baseline in Dice Similarity Coefficient (DSC) in \cite{zhou2021preservational}, leading to confusion about the validity of MG.

Another common experiment in related papers is to compare the performance of baseline and SSL methods using the varying percentage of labelled training data in the target task. We find that the significant performance gain from SSL pretraining only occurs when exploiting less than 50\% of the labelled data. When label utilization attains 100\% during fine-tuning, the performance improvement from SSL pretraining is basically less than 1\% unit \citep{tajbakhsh2019surrogate, tao2020revisiting, zhou2021preservational, taleb20203d} or even negative \citep{tao2020revisiting, taleb20203d}. This raises the suspicion that the benefits from SSL-based transfer learning in the small-scale labelled training dataset are due to the over-parametrization of standard models 
, as claimed in \cite{raghu2019transfusion} regarding the benefit of pretraining.

The aforementioned confusion can be attributed to inconsistent experimental setup and weak baselines. To make the results more convincing and offer a fair comparison, we establish a public code framework on PyTorch and reproduce all selected SSL methods across five target datasets. Due to the small scale of medical image datasets, we make full use of annotations in target tasks without any annotation dropping. Each method was trained \changed{M1.4.2}{\link{R1.4}}{three times} and the average results are presented in Table~\ref{tab:comparison}. Notably, our baselines consistently surpass previous implementations in \cite{zhou2021models,zhou2021preservational}, thus ensuring the performance reality and validity of different SSL approaches.

\textbf{Source code and configuration files are available at \url{https://https://github.com/EndoluminalSurgicalVision-IMR/Medical-SSL}. More details are provided in \ref{sec:appen_implementation}.}

\begin{table*}[!t] 
\renewcommand{\arraystretch}{0.9}
\setlength\tabcolsep{1.8pt}
\caption{Comparison of different methods on multiple target tasks. Here, we evaluate the Area Under the Curve (AUC) scores and accuracy (ACC) for NCC. For NCS, we report both the DSC and mean IoU (mIoU) with thresholds from 0.5 to 0.95. Note that ``mIoU'' stands for the average mIoU for the foreground class and background class (used in \cite{zhou2021models}), while ``mIoU+'' is only for the foreground class. For LCS, we report the DSC and IoU. For EPC, we compute the Kappa metric \citep{huang2021lesion} and ACC. For DVS, we select DSC and Sensitivity (SEN) for evaluation. The first two best results are bolded and underlined respectively. The statistical significance analysis of these methods is presented in \ref{sec:t_test}.} 
\centering
\subtable[3D tasks: pretraining on LUNA 2016]{
\resizebox{0.9\textwidth}{!}{
\begin{tabular}{ccccccccc}
	\toprule
		\multirow{2}{*}{Pretraining}    & \multirow{2}{*}{Method} & \multicolumn{2}{c}{NCC}           & \multicolumn{3}{c}{NCS}                         & \multicolumn{2}{c}{LCS}         \\ \cmidrule(r){3-4} \cmidrule(r){5-7}  
		\cmidrule(r){8-9} 
											&                         & AUC (\%)  & ACC (\%)      & DSC (\%)      & mIoU (\%)      & mIoU+ (\%)     & DSC (\%)      & IoU (\%)       \\ \toprule
											
		From scratch                     & Random init \citep{he2015delving}             & 98.53    & 97.00      & 73.79          & 80.86          & 62.58          & 93.83          & 88.50          \\ \midrule
		\multirow{1}{*}{Predictive SSL}  & ROT  \citep{taleb20203d}                   & 99.32    & \textbf{99.09}       & 73.25          & 80.64          & 62.15          & 94.49          & 89.65          \\ 
											& RPL   \citep{taleb20203d}                  & 99.29   & 97.87      & \underline{76.10}         & 81.36          & 63.51          & 94.86 & 90.18 \\ 
											& Jigsaw   \citep{taleb20203d}               & 98.64   & 98.81       &        75.23        &            81.90    &          64.66      & 94.36          & 89.42          \\ 
											& RKB  \citep{zhuang2019self}                  &\underline{99.41} & \underline{99.03}        & 74.22          & 80.88          & 62.61          & \underline{94.93}         & \underline{90.43}         \\ 
											& RKB+   \citep{zhu2020rubik}                 & 98.88  & 98.48    & 74.31         & 80.82         & 62.51         &   \textbf{95.46}            &     \textbf{91.41}          \\ \midrule
		\multirow{1}{*}{Generative SSL}  & AE                     & 97.76       & 97.62    & 74.26          & 81.15          & 63.17          & 93.77          & 88.92          \\ 
											& MG  \citep{zhou2021models}                    & 98.01     & 96.72     & 75.69          & \underline{82.13}         & \underline{65.08}          & 94.24          & 89.17          \\ 
											& PCRL \citep{zhou2021preservational}                   & 98.97       & 97.88    & 75.60          & 81.97          & 64.55          & 93.87          & 88.56          \\ \midrule
		\multirow{1}{*}{Contrastive SSL} & SimCLR \citep{chen2020simple}                 & 99.29    & 97.91      & 75.96          & 82.00          & 64.82          & 94.56          & 89.74        \\ 
											& BYOL \citep{grill2020bootstrap}                   & \textbf{99.52}& 97.65 & \textbf{76.13} & \textbf{82.27} & \textbf{65.37} & 94.43          & 89.56          \\ \bottomrule
		\end{tabular} }}
\\ 
\subtable[2D tasks: pretraining on EyePACS]{        \resizebox{0.65\textwidth}{!}{
       \begin{tabular}{cccccc}
	\toprule
		\multirow{2}{*}{Pretraining}    & \multirow{2}{*}{Method} & \multicolumn{2}{c}{EPC}           & \multicolumn{2}{c}{DVS}                           \\ \cmidrule(r){3-4} \cmidrule(r){5-6}  
											&                         & Kappa (\%)  & ACC (\%)      & DSC (\%)      & SEN(\%)      \\ \toprule
											
		From scratch                     & Random init \citep{he2015delving}             &76.32    & 77.33      & 78.57          & 75.45    \\ \midrule
		\multirow{1}{*}{Predictive SSL}  & ROT  \citep{taleb20203d}                   & 77.85    & 77.16       & 77.56          & 75.67     \\ 
											& RPL   \citep{taleb20203d}                  & 77.47   & 77.63      &79.21         & \underline{79.05}        \\ 
											& Jigsaw   \citep{taleb20203d}               & 78.29   & 76.76       &        78.72        &            77.89           \\ 
           \midrule
		\multirow{1}{*}{Generative SSL}  & AE                     & 70.28       & 74.33    & 79.28          & 77.23         \\ 
											& MG  \citep{zhou2021models}                    & 
           \textbf{78.66}     & \textbf{78.41}     & \underline{80.00}          & 78.63      \\ 
											& PCRL \citep{zhou2021preservational}                   & 77.92       & \underline{78.34}    & 79.49          & 76.93              \\ \midrule
		\multirow{1}{*}{Contrastive SSL} & SimCLR \citep{chen2020simple}                 & \underline{78.46}    & 77.91      & \textbf{80.75}          & \textbf{79.81}        \\ 
											& BYOL \citep{grill2020bootstrap}                   & 76.55& 77.58 & 78.42 & 77.33           \\ \bottomrule
		\end{tabular}} } 
      \label{tab:comparison}  
\end{table*}

\subsection{Preliminary analysis of different SSL methods}

Our reproduced results in Table~\ref{tab:comparison} first provide a fair comparison of all SSL types, including predictive, generative and contrastive SSL. Unlike results in previous SSL literature where their proposed SSL method can always prevail over others, we find that no single method is able to dominate all target datasets. This new finding indicates that there are a number of practical factors we have to consider for a specific target task, far beyond the design of a pretext task itself. In specific, these factors may include but are not limited to the characteristics of upstream and downstream medical data (i.e. class imbalance), effective implementations of SSL tasks, and useful combinations with other general training schemes. To investigate these issues, we conduct a fine-grained study on 3D datasets in the following four sections.

\section{Data Imbalance}
\label{sec:imbalance}

In clinical examinations, disease subjects are always much less than normal subjects, naturally leading to heavy ''data bias" in medical datasets. This poses a great challenge for deep learning models since typical supervised training based on imbalanced data annotations inevitably makes the model overfit to frequent classes and underperform in rare classes. Rather than relying on the imbalanced annotations, SSL pretraining could extract label-irrelevant but transferable features from the raw data for downstream classification. Due to this, a meaningful question arises:

\emph{How does SSL pretraining affect class-imbalanced learning?}

\begin{figure*}[!ht]
    \centering
    \includegraphics[width=0.8\linewidth]{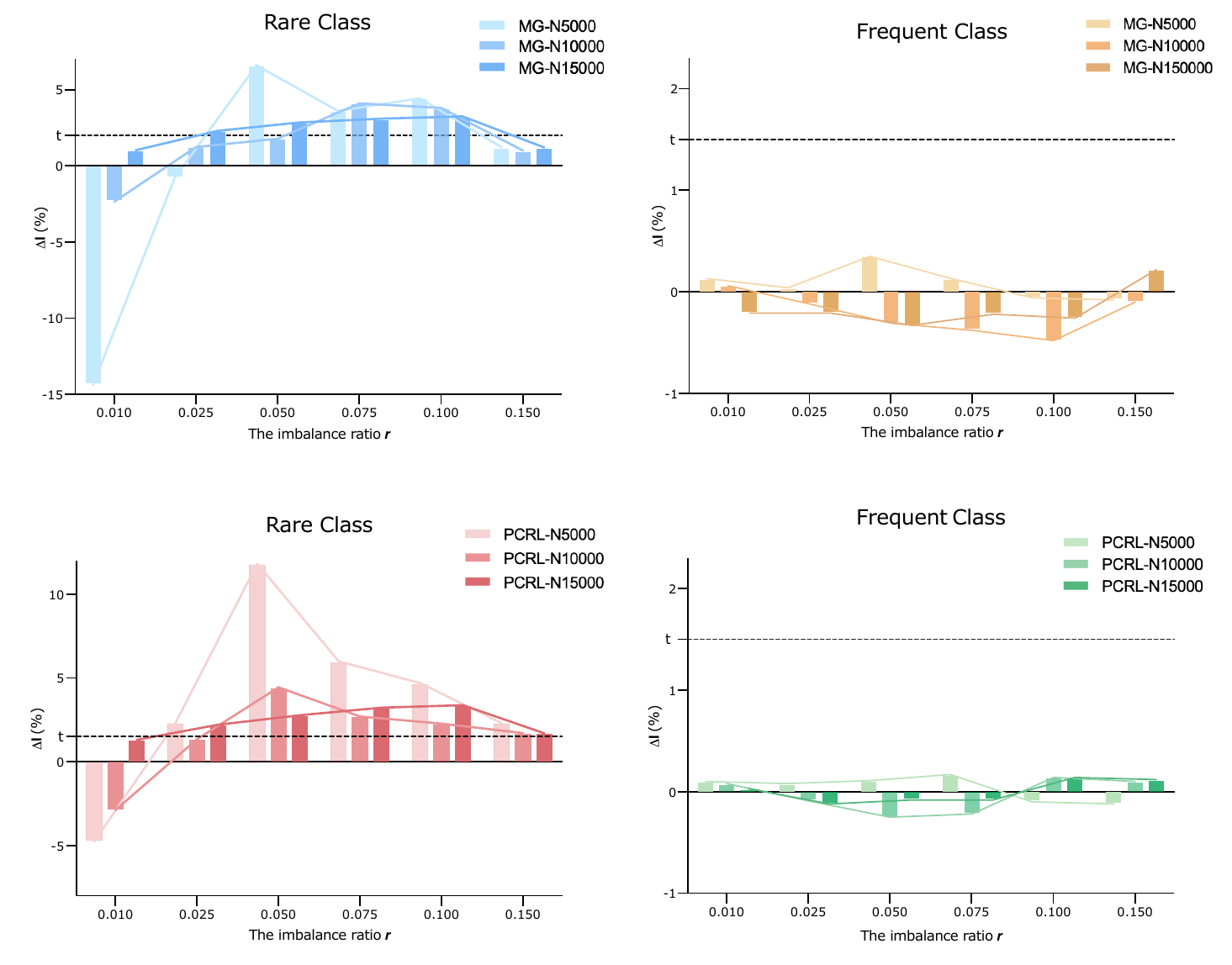}
    \caption{Class-imbalanced Classification. Each pretraining method is tagged as ``Method-N*''. Here, * is the size of a downstream training dataset, and $\Delta I$ is the error-rate gap between from-scratch models and SSL models. We set $t$ as a threshold representing a significant improvement. As the results show, the improvements from SSL methods are only significant in the rare (positive) class within a range of imbalance ratio rather than in the frequent (negative) class. }
    \label{fig:figure5}
\end{figure*}

\begin{figure*}[!ht]
    \centering
    \includegraphics[width=0.8\linewidth]{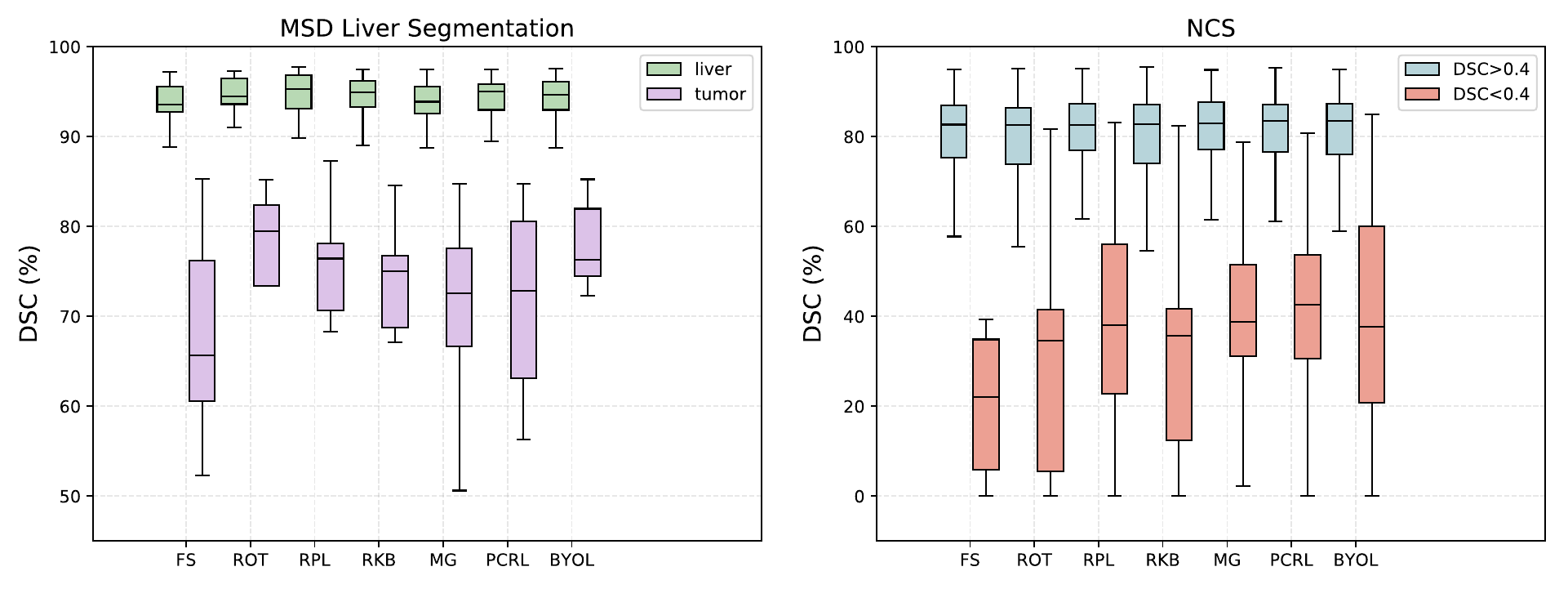}
    \caption{Class-imbalanced  Segmentation. We evaluate performance via DSC for the rare and frequent classes, respectively. Clear improvements from SSL methods are only noticeable for rare classes rather than frequent classes.}
    \label{fig:figure3}
\end{figure*}

In this section, we systematically investigate the effect of SSL algorithms on class-imbalanced downstream tasks. 
We first obtain basic intuitions on how class imbalance affects the downstream learning process through a simple theoretical model. Based on theoretical motivation, we design comprehensive experiments to confirm the theoretical insights. 
 \changed{M1.8.1}{
\link{R1.8}}{Main takeaways in this section are summarized below: \begin{enumerate}
\item The prior probability ratio related to both upstream and downstream data affects the target performance, namely the inherent data imbalance of pretraining data would impair downstream learning.~(Fig.~\ref{fig:figure5})
\item SSL methods improve the rare class more than the frequent class.~(Fig.~\ref{fig:figure5}, Fig.~\ref{fig:figure3})
\end{enumerate}}

\subsection{Theoretical Motivation}

In previous works, \cite{yang2020rethinking} has proved that semi-supervised training and self-supervised training can advance class-imbalanced learning. However, their theoretical discussion is based on the ideal assumption that the class priors are balanced, or that the underlying imbalance of unlabelled data does not affect the representation distributions. In this work, we further take into consideration the possible impact of biased priors and unlabelled data on the embedding distributions. To avoid the complexity of explaining neural networks, a simplified model is built to explore how SSL acts in pattern recognition.

Consider a simple binary classification problem with the data generating distribution $P_{XY}$ being a mixture of two Gaussians. Let label $Y = +1$ and $Y = -1$ refer to the rare class and major class, with a imbalance ratio of prior probabilities: $0 < \lambda=\frac{\mathbb{P}(Y = +1)}{\mathbb{P}(Y = -1)} < 1$. We consider the case where unlabelled data $\{\widetilde{X}_i\}, i=1,...,M$ (potentially also imbalanced) and labelled data $\{X_j, Y_j\}, j=1, ..., N$ from $P_{XY}$ constitute a pretraining dataset $D_{up}$ and a target dataset $D_{down}$, respectively. We assume that a self-supervised proxy task learns a representation $Z = \phi(X)$ from $D_{up}$. According to the superiority of SSL in previous works, it is reasonable to believe that the new feature $Z$ is better than the raw input $X$ for semantic recognition even though we might not explicitly know what the transformation $\phi$ is. Therefore, the target Bayesian decision boundary is determined by $Z$ on $D_{down}$.
Intuitively, the inherent similarity of samples can be learned from the visual data themselves rather than only semantic labels. It has been widely acknowledged that a good SSL model should map visually similar images close together in the embedding space, even though the class labels are unseen \citep{wu2018unsupervised, saunshi2019theoretical, liu2021self}.
With that said, we can assume that $Z$ conditioned on $Y = +1$ and $Y = -1$ follows different Gaussian distributions. Formally, $Z|Y=+1 \sim N(\mu_1, \sigma^2)$ and $Z|Y=-1 \sim N(\mu_2, \sigma^2)$. For simplicity, we ignore the variance difference between the two classes. Assuming $\mu_1 > \mu_2$, the optimal Bayesian classifier can be expressed as $f(Z) = sign(Z-\frac{\mu_1+\mu_2}{2}+\frac{\sigma^2\ln(\lambda)}{\mu_1-\mu_2})$ \changed{M1.5.a}{\link{R1.5.a}}{\citep{theodoridis2006pattern}}, i.e. $X$ is classified as $Y = +1$ when $f(\phi(X)) > 0$. Our estimation of $f$ is composed of two terms: $\theta_1 = \frac{\mu_1+\mu_2}{2}$ and $\theta_2 = \frac{\sigma^2\ln(\lambda)}{\mu_1-\mu_2}$. \changed{M1.5.b}{\link{R1.5.b}}{The estimation of the former term is easily obtained by the maximum likelihood estimate of the mean for Gaussian densities  \citep{theodoridis2006pattern}}: $\hat{\theta}_1=\frac{\sum_{k=1}^{N^+}Z_k^+/N^++\sum_{k=1}^{N^-}Z_k^-/N^-}{2}$, where $N^+$ and $N^-$ are the number of positive class and negative class in $D_{down}$ respectively. According to the Gaussian concentration inequality, we have:

\textbf{Theorem 1.} \emph{Consider the above setup. For any $t > 0$, with probability at least $1-2e^{\frac{-2t^2}{\sigma^2}\frac{N^+N^-}{N^++N^-}}$ our estimated $\hat{\theta}_1$ satisfies:}
\begin{equation} 
 |\hat{\theta}_1-\frac{\mu_1+\mu_2}{2}| \leq t
\end{equation}

\textbf{Interpretation 1.} The proof is provided in \ref{sec:proof_theorem1}. We discuss the estimation of $\theta_1$ and $\theta_2$ separately: (1) Theorem 1 indicates that the imbalance degree of labelled training data affects the chance of obtaining a good estimate of $\theta_1$ since that $\frac{N^+N^-}{N^++N^-}$ is maximized when $N^+ = N^-$. (2) The second term of $\theta_2$ implies that extreme data imbalance introduces large estimation uncertainty. As all the training samples are derived from $P_{XY}$ in our prerequisite, class prior information is relevant to the unlabelled data as well. Therefore, it is inaccurate to estimate $\lambda$ with $\frac{N^+}{N^-}$ especially when the scale of $D_{up}$ is larger than $D_{down}$, leading to non-negligible estimation uncertainty. For a small $\lambda$ ( $0 < \lambda < 1$ ), $|\ln(\lambda)|$ turns significant so that the term $\theta_2 = \frac{\sigma^2\ln(\lambda)}{\mu_1-\mu_2}$ dominates the decision function $f$, thereby magnifying the estimation uncertainty. 

We then consider a large $\lambda$, i.e. there is less class imbalance in upstream and downstream data. Based on the understanding in Interpretation 1, it still have a high probability to get an accurate estimate of the optimal classifier. We further seek the error rate of such a good classifier. Suppose that the Bayesian decision boundary splits the whole feature space into two parts: $\Gamma_+$ and $\Gamma_-$, the error probability of each class can be computed as: $\epsilon_+ = \int_{\Gamma_-}p(Z|Y=+1)dZ$, $\epsilon_- = \int_{\Gamma_+}p(Z|Y=-1)dZ$. Then, we can derive the \textit{Chernoff} error upper bound according to \cite{fukunaga2013introduction}:

\textbf{Theorem 2.} \emph{Consider the above setup. Given the ratio of prior probabilities $\lambda$ and the Bhattacharyya distance of two classes $D_B=\frac{1}{8}\frac{(\mu_2-\mu_1)^2}{\sigma^2}$, the error probability of each class of Bayes's classifier satisfies:}

\begin{equation}       
	\left\{              
	  \begin{array}{l}   
		\epsilon_+ \le \frac{1}{\sqrt{\lambda}}e^{-D_B}\\ 
		\epsilon_- \le \sqrt{\lambda}e^{-D_B}\\
	  \end{array}
	\right.                 
	\end{equation}

\textbf{Interpretation 2.} The proof is provided in \ref{sec:proof_theorem2}. Theorem 2 reveals several interesting aspects: 
(1) The prior probability ratio $\lambda$ is crucial for the performance of the classifier. (2) SSL methods improve the rare class more than the major class. For a supervised learning method, the raw input $X$ is directly used for Bayes's decision. Let $X|Y = +1 \sim N(\mu_1', \sigma'^2)$ and $X|Y = -1 \sim N(\mu'_2, \sigma'^2)$, it is easy to verify the corresponding upper bound $B$ of error rate: $B(\epsilon'_+) = \frac{1}{\sqrt{\lambda}}e^{-D_B'}$, $B(\epsilon'_-) = \sqrt{\lambda}e^{-D_B'}$. 
\changed{M1.5.f, M2.7}{\link{R1.5.f}\link{R2.7}}{Previous observations on medical datasets have shown that the proposed SSL method enables a better class-separated embedding space than the randomly initialized model \cite{feng2020parts2whole}
or the supervised counterpart \cite{truong2021transferable}. Besides, we find that several SSL methods offer higher class separation than the baseline in \ref{appen:t_sne}. Therefore, we assume that good SSL representations are supposed to have a larger inter-class gap than the raw input:}
$D_B > D_B'$. Therefore, we have $\Delta=e^{-D_B'} - e^{-D_B} > 0$. Apparently, $\frac{1}{\sqrt{\lambda}} > \sqrt{\lambda}$ when $0 < \lambda < 1$. Therefore, the positive error rate is reduced much more than the negative error rate with SSL methods. Precisely, $\Delta_+ = B(\epsilon'_+) - B(\epsilon_+) > \Delta_-= B(\epsilon'_-)-B(\epsilon_-)$.

\begin{figure}[!ht]
    \centering
    \includegraphics[width=0.8\linewidth]{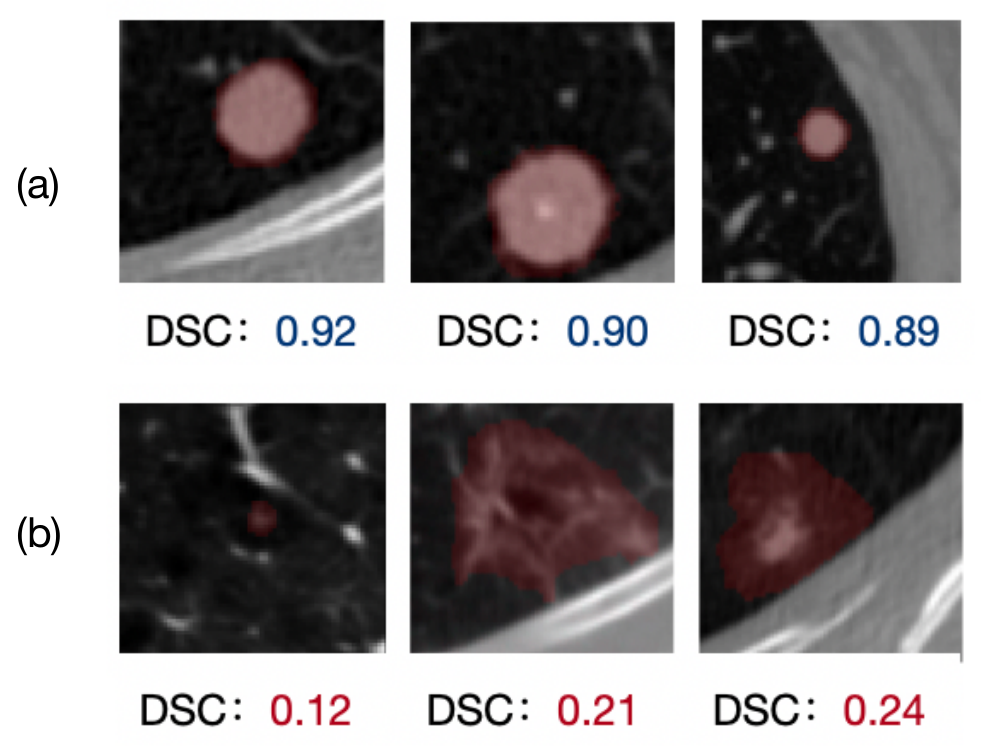}
    \caption{Lung nodules with predicted DSC in NCS. (a) and (b) display samples with large and small DSC, respectively. The ground truth of each nodule is highlighted in red. We observe that solitary solid nodules often get accurate predictions, while subsolid nodules such as ground-glass or part-solid nodules are difficult to segment. The solid nodules in (a) always present isolated, circular and high-density shadows, which are homogeneous; The subsolid nodules in (b) exhibit mutable forms, which are heterogeneous. Hence, we regard the former as a uniform and frequent class, while the latter is assigned to rare classes.}
    \label{fig:lung_nodules}
\end{figure}

\subsection{Results}

Inspired by the theoretical analysis above, we conduct detailed experiments to explore the effect of SSL on class-imbalanced learning in medical image classification and segmentation. 

\textbf{Classification} 
We present an imbalanced classification on NCC dataset. The pretraining data is always fixed, while the target data is sampled differently for exploring the effect of SSL on class-imbalanced learning. Let $x$ and $y$ denote the input and corresponding label, respectively. In NCC, $x$ with $y=1$ is a true-positive nodule and $x$ with $y=0$ is a false-positive nodule, where the amount of the former is much less than the latter. Following \cite{cao2019learning}, we define a variable $r$ as the ratio between the size of the rarest class and the most frequent class. Then we construct distributions with varying imbalance ratio $r$ in $\{0.025, 0.05, 0.075, 0.1, 0.15\}$ for downstream tasks. For each imbalance ratio, we further adjust the number of training samples $N$ to $\{5000, 10000, 150000\}$. Note that we fix the test set for different dataset variants. Apparently, the performance of a baseline model (from scratch) is not only related to $r$ but also $N$. Given pretrained SSL models, we fine-tune these models with different downstream training datasets and compare each performance with the corresponding baseline model. Fig.~\ref{fig:figure5} shows how the performance improvement from SSL pretraining $\Delta I =\Phi_{SSL}-\Phi_{FS}$ varies over different imbalanced distributions, where $\Phi=1-\epsilon$ (the error rate). In clinical applications, missing true positive cases is much more costly than false positive cases. Hence, we report the $\Phi$ of each class separately and focus on the rare class. We surprisingly observed that SSL pretraining performs poorly (even worse than baseline) on extremely imbalanced and relatively balanced cases. Our main observations are concluded as\changed{M1.6}{\link{R1.6}}:

\begin{enumerate}[label = (\arabic*)]
\item SSL yields a marginal impact on the frequent class (less than 0.5\%) compared to the rare class, so the rest of our analysis focuses on the rare class. 
\item SSL achieves very limited improvement on relatively balanced labelled data with a large $r$.
\item SSL would impair model performance on extremely imbalanced labelled data with a small $r$.
\end{enumerate}

These observations indeed support our theoretical insights. In specific, both (1) and (2) highlight that the benefits of SSL are mainly reflected in the improvement over imbalanced categories, further confirming Interpretation 2. In contrast, the impact of SSL on the frequent class is negligible. The extreme case in (3) is relevant to Interpretation 1, which declares that the inherent data imbalance of pretraining medical data also has a negative effect on downstream learning, from the perspective that both unlabelled and labelled data naturally obey the same class priors. Here, based on broad experiment results, we attempt to explain the negative effect of SSL under extreme imbalance as: In the SSL pretraining stage, biased training data results in skewed representation learning oriented to frequent samples. This representation offset would be expected to get corrected by downstream supervised fine-tuning but is instead amplified when the downstream labelled data is severely biased and it fails to provide profitable supervision signals. Therefore, the cumulative biased representation learning under extreme data imbalance leads to poor performance in the rare category. Except for this extreme case, the labels of the downstream data will be conducive to representation learning. At this point, the skewed SSL representations can be easily rectified by downstream strong supervision. 


\textbf{Segmentation} We fine-tune various SSL pretrained models on NCS and MSD datasets, and compare each performance with the corresponding baseline model. In MSD, the liver and tumor can be regarded as the frequent class and rare class, respectively. 
\changed{M2.6.1}{\link{R2.6}}{The imbalance ratio of tumor and liver regions is nearly 0.047.} \changed{M1.7}{\link{R1.7}}{In NCS, \cite{kaluva2020automated} has shown that solid nodules are more frequent than subsolid nodules, resulting in the problem of data imbalance regarding  nodule types. Studies in \cite{li2019detectability, kaluva2020automated, farhangi20173} suggest that the subsolid nodules tend to be hard samples for detection and segmentation algorithms. In our experiments, it can be observed that the predictions of NCS always fall into two extremes, either with very high dice score or very low dice score. As displayed in Fig.~\ref{fig:lung_nodules}, subsolid nodules are faced with poor segmentation performance due to heterogeneous textures. Based on the closeness between the hardness of predicting subsolid nodules and their rarity, we treat samples with dice values less than 0.4 as hard samples in the rare nodule category.} 
\changed{M2.6.2}{\link{R2.6}}{The imbalance ratio of hard and simple cases is around 0.087.} Fig.~\ref{fig:figure3} suggests the same trend as the classification task that SSL tends to exhibit larger gains on the rare class than the frequent class across two segmentation datasets. 
\begin{figure*}[!th]
    \centering
    \includegraphics[width=0.8\linewidth]{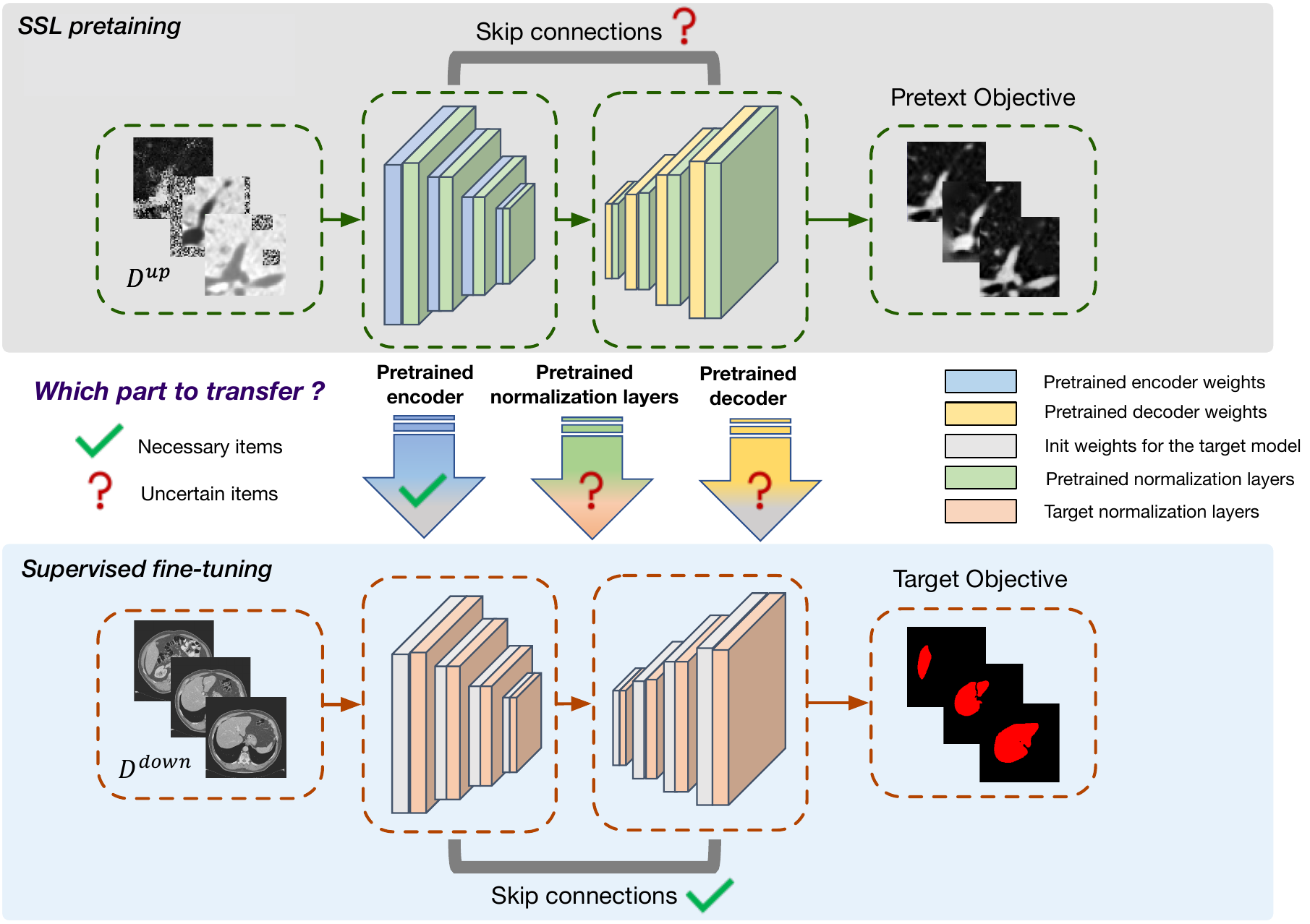}
    \caption{Different modules in U-shape network to be explored during transfer learning. We illustrate several architecture-related issues during the fine-tuning stage. $D_{up}$ and $D_{down}$ represent the upstream dataset and the downstream dataset, respectively. The remarkable domain difference between $D_{up}$ and $D_{down}$, and the objective difference between the pretext task and the target task both suggest the demand for effective weight transfer schemes. }
    \label{fig:network_arch}
\end{figure*}
\begin{table*}[!h]
	\renewcommand{\arraystretch}{0.7}
	\caption{The effect of each module in the U-shape architecture. For each reconstruction-based SSL method, we first evaluate the performance in target tasks when using separately pretrained encoder, decoder and both. Note that we use DSC as the performance metric for both NCS and LCS, and AUC score for NCC. Then, we re-train each SSL model without the skip connections in the U-shape architecture and report its downstream performance. }
	\centering
 \resizebox{0.8\textwidth}{!}{
	\begin{tabular}{cccccccc}
	\toprule
	\multirow{2}{*}{Model}         & \multicolumn{3}{c}{Architecture} & Pretext task            & \multicolumn{3}{c}{Target task}                                   \\ \cmidrule(r){2-4}\cmidrule(r){5-5}\cmidrule(r){6-8}
									& w/o encoder    & w/o decoder    & w/o skip    & MAE                     & \multicolumn{1}{c}{NCS}   & \multicolumn{1}{c}{LCS}     & NCC   \\ \toprule
	\multirow{4}{*}{Model Genesis} &           &          &        & \multirow{3}{*}{0.0053} & \multicolumn{1}{c}{75.69} & \multicolumn{1}{c}{94.24} & 98.01 \\
									&       &   \checkmark       &       &                         & \multicolumn{1}{c}{75.70} & \multicolumn{1}{c}{94.08}   & —     \\
									&   \checkmark     &         &        &                         & \multicolumn{1}{c}{74.77} & \multicolumn{1}{c}{93.60}   & —     \\ 
									&          &         &   \checkmark      & 0.0095                  & \multicolumn{1}{c}{77.09} & \multicolumn{1}{c}{94.72}   & 99.00 \\ \midrule
	\multirow{4}{*}{PCRL}          &        &    &       & \multirow{3}{*}{0.0402} & \multicolumn{1}{c}{75.60} & \multicolumn{1}{c}{93.87}   & 98.97 \\
									&         &  \checkmark       &       &                         & \multicolumn{1}{c}{74.02} & \multicolumn{1}{c}{93.25}   & —     \\
									&      \checkmark      &          &      &                         & \multicolumn{1}{c}{74.43} & \multicolumn{1}{c}{93.56}   & —     \\ 
									&        &           &   \checkmark       & 0.0431                  & \multicolumn{1}{c}{74.30} & \multicolumn{1}{c}{93.18}   & 98.34 \\ \bottomrule
	\end{tabular}}
	\label{tab:table2}
	\end{table*}
\section{Network Architecture}
\label{sec:u-shape}

\begin{figure*}[t]
    \centering
\includegraphics[width=0.9\linewidth]{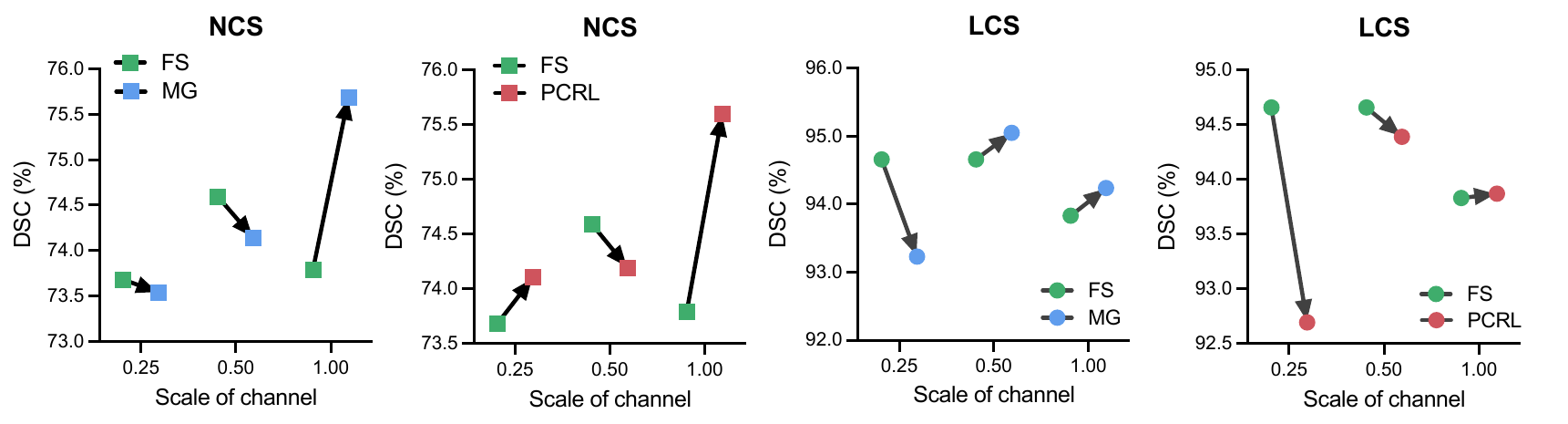}
    \caption{The performance of SSL pretraining in NCS and LCS when scaling down the channels of U-Net. The channels of convolution blocks in the original U-Net are 64-128-256-512-256-128-64, and the scale $s$ means using $s\cdot$(64-128-256-512-256-128-64) channels. FS stands for training from scratch.}
    \label{fig:UNet_channel}
\end{figure*}

The most prevalent architecture for medical image segmentation is the U-shape network ~\citep{ronneberger2015u, cciccek20163d}, consisting of a down-sampling encoder, a symmetric up-sampling decoder and skip connections. As a pixel-wise classification task, segmentation relies on fine-grained features to localize the region of object~\citep{hariharan2015hypercolumns}. To integrate features at different levels for precise segmentation, skip connections are proposed to bridge the encoder and decoder, which has achieved outstanding performance \citep{drozdzal2016importance}. Constrained by the ``pretrain-then-finetune" paradigm for transfer learning, it is necessary to ensure that self-training adopts the same encoder architecture as downstream tasks. As illustrated in Fig.~\ref{fig:method_overview}, the network head after the encoder varies for different proxy and target tasks. Apart from the indispensable encoder reuse, the transferring of the pretrained decoder's parameters has also been discussed in previous papers. For instance, \cite{tao2020revisiting} claimed that the decoder pretraining alleviates the negative influence of randomly-initialized decoder in the medical segmentation task. However, this claim remains unsubstantiated, leading to confusion about the up-to-downstream adaptation schemes. For effective SSL implementations, we investigate the influence of different modules in the U-shape architecture as well as the fine-tuning strategy during transfer learning, as depicted in Fig.~\ref{fig:network_arch}. Moreover, we evaluate the benefit of SSL pretraining when scaling down the network capability to reveal the essence of SSL.
 \changed{M1.8.2}{
\link{R1.8}}{Main takeaways in this section are summarized below: \begin{enumerate}
\item The decoder risks overfitting to the reconstruction task, thus offering little benefits for downstream tasks. To mitigate this issue, removing the skip connections in pretraining could lead to better representations.~(Table~\ref{tab:table2}, Fig.~\ref{fig:figure6})
\item The benefit of SSL pretraining may come
from an alleviation of over-parameterization~(Fig.~\ref{fig:UNet_channel})
\item It is recommended to recollect the
target BN statistics for inference as source-target mixed BN statistics usually induce performance degradation under a large domain gap.~(Fig.~\ref{fig:BN})
\item Full fine-tuning is more advantageous than warm-up fine-tuning due to the essential gap between SSL proxy tasks and downstream tasks.~(Fig.~\ref{fig:fine_tuning})
\end{enumerate}}

\subsection{Different modules in the U-shape architecture}
\label{sec:u_shape_arch}

To enable the reuse of pretrained parameters, reconstruction-based SSL methods always adopt the same U-shape architecture as medical segmentation tasks. However, we argue that the U-shape architecture with skip connections that are particularly designed for segmentation might be sub-optimal for generative pretext tasks. As the encoder and decoder serve different functions in solving a task, we are also concerned about the roles they play when transferred to target tasks. In specific, different parts of the model have different impacts on downstream tasks, and understanding these effects will facilitate the selection of which part to transfer. Thus far, the role of each component in the network architecture is still less explored. In this section, we systematically analyze the impact of encoder, decoder and skip connections on transfer learning by thorough experiments. Accordingly, we transfer different components of models pretrained by generative pretext tasks to various downstream tasks and report the performance in Table~\ref{tab:table2}. In addition, we display the reconstruction error of SSL models on the validation set as an indicator of the difficulty of a pretext task.

\textbf{The decoder risks over-fitting to the reconstruction task, thus offering little benefits for downstream tasks.} In MG, the model with a randomly-initialized decoder achieves comparable performance to the pretrained decoder, while the absence of the pretrained encoder results in noticeable performance degradation. On the one hand, this overturns the declaration in the previous work that the pretraining decoder provides a huge boost on target performance \citep{tao2020revisiting}. We argue that the advantages of the SSL task might come from valuable anatomical information learned by the encoder rather than pixel-wise details in the decoder. 
Nonetheless, PCRL avoids overfitting to visual details by introducing contrastive learning and transformation-conditioned reconstruction into the SSL framework, thereby providing a more transferable decoder. On the other hand, the poor performance of only transferring the decoder proves that it is not feasible to initialize parameters in a truncated manner of randomly initializing previous layers (encoder) and transferring pretrained parameters after these layers (decoder). To ensure the validity of the self-supervised learned representations, the transferred parameters should not be disconnected.
\begin{figure}[!ht]
	\centering
	\includegraphics[width=0.9\linewidth]{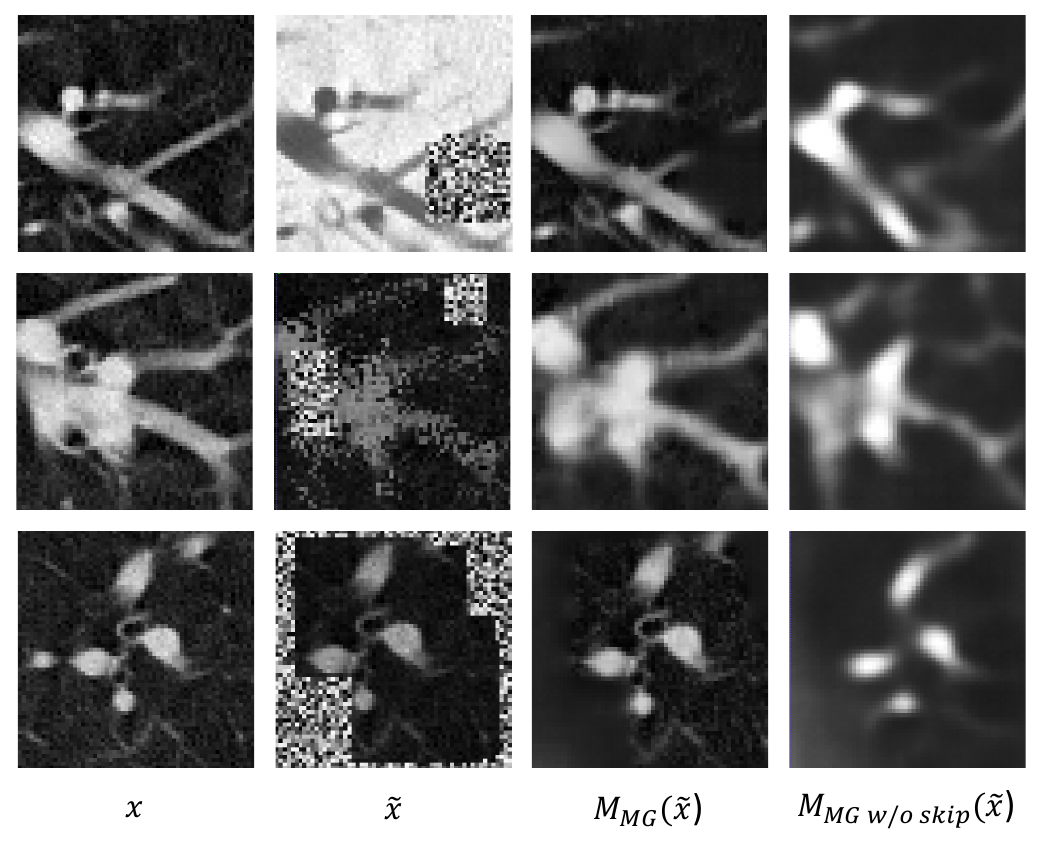}
	\caption{Restored images from generative SSL models. From left to right, the original images, the input images deformed via MG method, the restored images by the typical MG model and the reconstructed images by the MG model without skip connections.}
	\label{fig:figure6}
\end{figure}

\textbf{Removing the skip connections in pretraining could mitigate the over-fitting of the network to reconstruction details.} 
Although the skip connections in the U-shape network lead to more precise predictions via assembling high-resolution and low-level features, they can hurt the model generalization for reconstruction proxy tasks. 
As shown in Fig.~\ref{fig:figure6}, the restored images from the MG model retain almost all the subtle details of volumes, exhibiting a considerably high reconstruction quality. Then, we remove the skip connections in the U-shape network for MG and find that such a model only restores the major structures. It is therefore reasonable to attribute these minor details in MG-restored images to the skip connections. However, what really matters is the representative features contributing to subsequent semantic tasks rather than the performance of the pretext task itself. If most of the information can be easily transmitted from the front layers to the decoder, the encoder will have little incentive to encode valuable features for the decoder. Instead, removing the skip connections when solving pretext tasks would force the network to extract high-level anatomical features. The results in Table~\ref{tab:table2} indicate that removing the skip connections in MG achieves worse reconstruction quality on the pretext task, but better performance on downstream tasks. Nevertheless, removing the skip connections is detrimental when the pretext task is hard enough, e.g. PCRL. In specific applications, it is necessary to consider the difficulty of a pretext task to decide whether to address the potential overfitting issue.

\subsection{Network capability}
\changed{M2.2}{\link{R2.2}}{Inspired by \cite{raghu2019transfusion}, we investigate the benefit of SSL when scaling down the capability of the U-shape network. The results in Fig.~\ref{fig:UNet_channel} show that the gain from SSL decays in U-Net with reduced channels. On the one hand, the reason can be ascribed to the deteriorated representation quality in pretrained models with lower capability. On the other hand, 
we notice that the 0.5$\times$ scaled U-Net trained from scratch instead outperforms the original U-Net in both NCS and LCS. This phenomenon corroborates the finding in \cite{raghu2019transfusion} that the benefit of pretraining may come from an alleviation of over-parameterization}.

\subsection{Normalization Layer}

\begin{figure*}[!ht]
    \centering
    \includegraphics[width=0.9\linewidth]{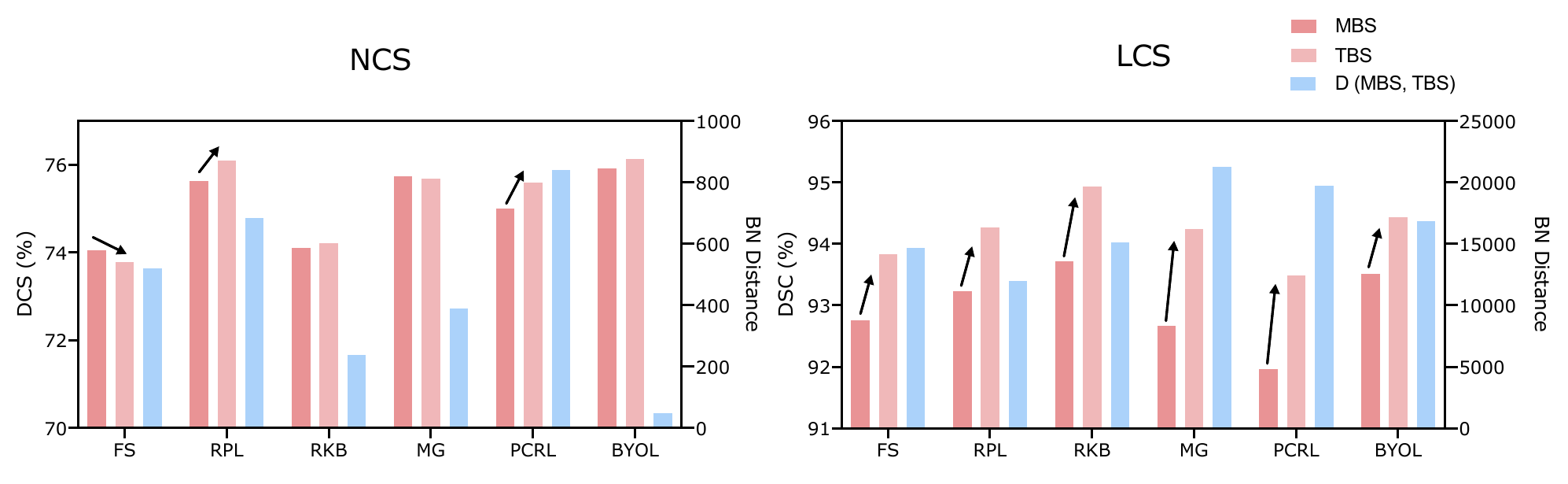}
    \caption{Comparison of different BN statistics. For each method, the predictions are obtained by Mixed BN statistics (MBS) and Target BN Statistics (TBS), respectively. We also report the BN distances of TBS and MBS.}
    \label{fig:BN}
\end{figure*}

We notice that current SSL studies have neglected an important issue that the transfer might be hurdled by the domain shift between proxy and downstream tasks.
It has been proved that the domain traits are included in Batch Normalization (BN) statistics \citep{li2016revisiting}. In the field of domain adaptation, the importance of separating BN layers for different domains has been emphasized frequently \citep{chang2019domain}. Motivated by this, we attempt to seek an appropriate BN adaptation policy for SSL-based transfer learning.

During the standard fine-tuning stage, the statistics in BN layers merge the target data with the historical data used in pretraining.
The mixed data distribution would be detrimental to the target task performance if there exists a significant source-target domain gap. Thus, we recommend re-collect the normalization statistics of the target data for better inference.
Following \cite{zhou2022generalizable}, we compute the distance between the Mixed and Target BN statistics (MBS and TBS in short) as a quantitative measurement of distribution shift:\\
\begin{equation}   
D(MBS, TBS) = \sum\nolimits_{l=1}^L \Vert (\mu_{mixed}^l-\mu_{target}^l) \Vert_2^2  + \Vert ({\sigma^l}_{mixed}^2-{\sigma^l}_{target}^2) \Vert_2^2      
\end{equation}
where $(\mu^l_{mixed}, \sigma^l_{mixed})$ and $(\mu^l_{target}, \sigma^l_{target})$ are the corresponding statistic values in MBS and TBS. $\{1, 2, ..., L\}$ refer to the index of encoder layers.

\textbf{Mixed BN statistics usually induce performance degeneration at a large distribution distance so that recollecting the target BN statistics for inference is necessary.}  Fig.~\ref{fig:BN} compares the performance of MBS and TBS in two segmentation tasks. For NCS, the same upstream and downstream data means a minimal domain gap. Still, the representation gap between proxy and target tasks or the data gap between training and test data might account for the slight performance difference between MBS and TBS. In most of the SSL methods, MBS and TBS do not clearly outperform the other. For transfer learning in LCS with a large domain gap between the upstream chest CT scans and target abdominal CT scans, TBS outperforms MBS by visible gaps. Therefore, 
we believe that using TBS for inference is a reliable way to avoid the potential risk of mismatched BN statistics.

\subsection{Fine-tuning strategy}

\begin{figure*}[!ht]
    \centering
    \includegraphics[width=1\linewidth]{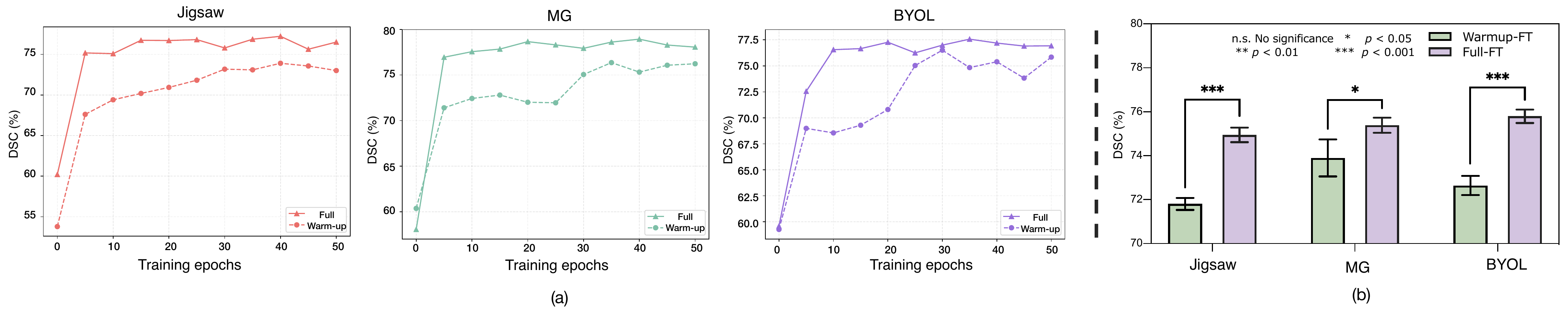}
    \caption{Results of different fine-tuning schemes in (a) the validation set of NCS and (b) the test set of NCS}
    \label{fig:fine_tuning}
\end{figure*}

Given an SSL pretrained model, how to reuse the parameters determines whether meaningful features are transferred well for solving downstream tasks. Throughout related works, there are mainly three choices for fine-tuning the pretrained model to target tasks: fixed \citep{goyal2019scaling}, warm-up \citep{taleb20203d}, and full fine-tuning\citep{zhou2021models}. The fixed way is to freeze the encoder and fine-tune the network head (decoder stacks or fully-connected layers) for the target task. The full way is to fine-tune all the weights with a relatively small learning rate. As a neutral to the above two ways, the warm-up way is to first fix the encoder for a few epochs and then execute the full fine-tuning. We first exclude the fixed way as fixed pretrained weights are hardly able to perform well in the target task, especially when there exists a distinct source-target domain gap. Then, we investigate the warm-up and full ways for transfer learning on three representative pretext tasks of predictive, generative and contrastive SSL: Jigsaw, MG and BYOL.

\textbf{Full fine-tuning is more advantageous than warm-up fine-tuning due to the essential gap between self-training and downstream semantic tasks.} Our experiments in Fig.~\ref{fig:fine_tuning} present a consistent outcome that the full fine-tuning is superior to the warm-up fine-tuning among three SSL tasks. 
It is worth mentioning that the warm-up fine-tuning causes the smallest performance decline in the generative model, probably because the reconstruction task is closer to the segmentation task intrinsically (both more concerned with fine-grained features). Therefore, we recommend full fine-tuning with a small learning rate instead of fixed pretrained parameters for SSL models.

\section{Pretext-target task relatedness}
\label{sec:pretexttasks}

With the rapid development of SSL, more and more interesting pretext tasks are being developed. These works focus on presenting a new proxy task and proving its priority over other methods, but they lack more detailed comparisons, such as the effective scope of each SSL method. Actually, our reproduction of diverse SSL methods in Table~\ref{tab:comparison} has shown no clear winner across all target tasks, indicating that universal SSL pretraining remains infeasible. Thus, researchers have to confront a fundamental question below:

\emph{How to implement effective SSL pretraining from various SSL approaches to facilitate a specific target task?}

\begin{figure*}[!ht]
    \centering
    \includegraphics[width=0.85\linewidth]{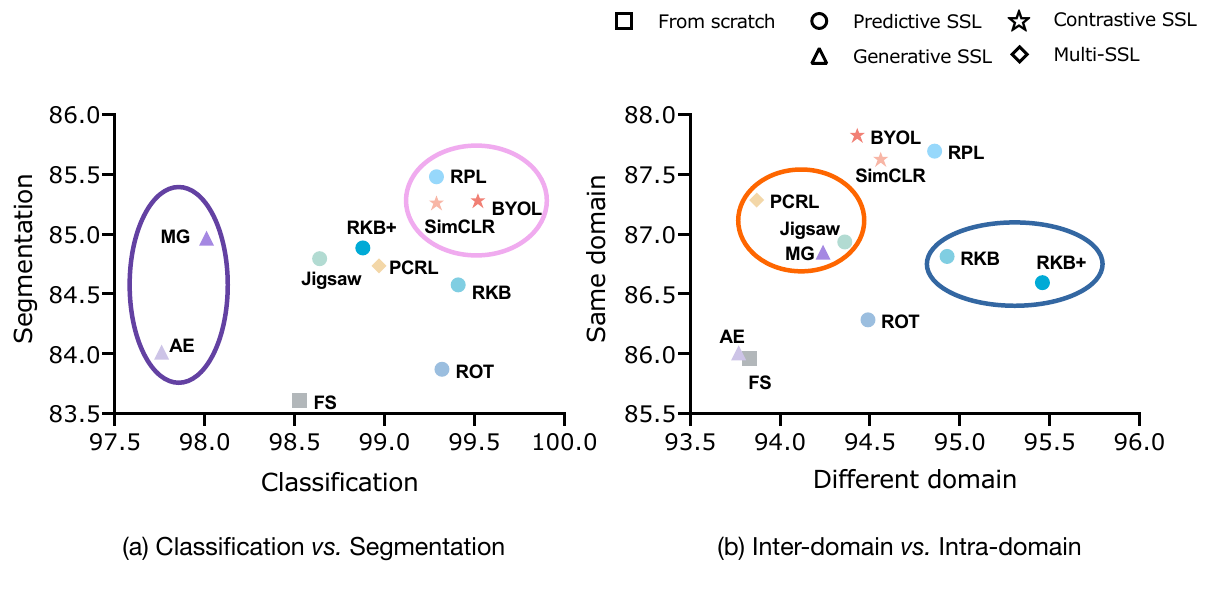}
    \caption{Comparison of different methods in terms of target tasks and the up-downstream data domains. (a) shows the performance of methods in regular medical recognition tasks, namely classification and segmentation tasks. Here, we treat the average DSC of NCS and LCS as the measurement of the segmentation task. (b) draws the behaviours of methods in inter-domain and intra-domain transfer learning. Since the NCC and NCS data are both from the same domain in the pretraining dataset, we treat the average value of NCC's AUC score and NCS's dice score as the measurement of intra-domain transfer. As for the inter-domain transfer, the DSC is measured in LCS. }
    \label{fig:comparison}
\end{figure*}

In this section, we seek to answer this question in three steps. In the first step, we analyze the adaptation performance from different types of pretext tasks to downstream tasks and summarize several guidelines for selecting an appropriate SSL task. Then, we discuss the influence of data processing on implementing predictive SSL. Finally, we conduct a feature-level comparative analysis on diverse SSL tasks to explain their differing performance on target tasks.  \changed{M1.8.3}{
\link{R1.8}}{Main takeaways in this section are summarized below: \begin{enumerate}
\item  It is necessary to select appropriate SSL tasks according to specific concerns of a target task, e.g. generative SSL underperforms in the classification
task. For blindly choosing, contrastive SSL and RPL could be the go-to solutions for researchers.~(Fig.~\ref{fig:comparison})
\item The accuracy of proxy tasks is not positively correlated to the performance of target tasks.~(Fig.~\ref{fig:input_size})
\item Predictive SSL models focus more on the object of interest itself while generative SSL models pay attention to the surroundings.~(Fig.~\ref{fig:CAM})
\item Higher feature reuse of pretrained features is more productive for transfer learning.~(Table~\ref{tab:CKA})
\end{enumerate}}
\subsection{Selection of Pretext Tasks}
For a specific target task, we put forward several considerations when selecting an SSL pretext task. The followings are our discussions and suggestions:

\textbf{The generative SSL underperforms in the classification task.} As drawn in Fig.~\ref{fig:comparison}(a), the generative SSL lags far behind other-type SSL in classification. A possible explanation may be the different attention of reconstruction and semantic classification, with the former focusing more on pixel-wise information and the latter preferring high-level representations. In contrast, the predictive and contrastive SSL methods pay attention to the high-level anatomical structures. Therefore, we recommend predictive and contrastive SSL pretraining over generative SSL pretraining for the classification task.

\textbf{Take into account properties of the target task when selecting the pretext task for pretraining.} As a prominent example, predicting rotations is not a viable auxiliary task for lung nodule segmentation. Since circular lung nodules tend to be rotation invariant, rotation-relevant features should not occupy too much attention of the network. Consequently, rotation-related SSL models consistently perform poorly in NCS. The inferior performance of RKB to Jigsaw also confirms this idea since an additional rotation prediction sub-task is added to RKB compared with Jigsaw. 

\textbf{For the adaptation under a large domain shift between upstream and downstream data (e.g. LCS), it is important to prevent potential overfitting to the upstream data in predictive or generative SSL.} In Fig.~\ref{fig:comparison}(b), we can observe that few predictive SSL and generative SSL perform poorly in intra-domain transfer learning. In the predictive SSL, the Jigsaw model might learn previously determined orders instead of general features. In contrast, harder predictive tasks tend to perform better, such as predicting the relative position of two randomly cropped patches or integrating more orthogonal sub-tasks to Jigsaw (RKB, RKB++). In generative SSL, previous reconstruction tasks based on the typical U-Net backbone are more likely to learn trivial solutions, leading to a limited transferable performance in LCS. Removing the skip connection would mitigate this issue. 

\textbf{For blindly choosing, contrastive SSL and RPL could be the go-to solutions for researchers.} From Fig.~\ref{fig:comparison} above, we can see that SimCLR, BYOL, and RPL robustly achieve good performance across all target tasks. This reveals promising advantages of contrastive SSL for learning more robust representations in medical imaging compared with other types of SSL.

\subsection{Sensitivity of predictive SSL methods with respect to the input size}
\label{sec:input_size}

\begin{table}[!t]
	\caption{The input size of different tasks. In the upstream tasks, the number after ``ROT" and ``RPL" stands for the input size in the axial (H-W) view. In the downstream tasks, the 0.5 in LCS means that we apply a 0.5x downsampling to the inputs due to limited GPU memory.}
	\centering
	\renewcommand{\arraystretch}{0.9}
  \resizebox{0.4\textwidth}{!}{
	\begin{tabular}{ccc}	
	\toprule
	&Task                            & Input size (H$\times$W$\times$D)     \\\toprule
	Upstream                                                 & ROT-48  & 48$\times$48$\times$32             \\
																					& ROT-64  & 64$\times$64$\times$64             \\
																					& ROT-128 & 128$\times$128$\times$64           \\
																		
																					& RPL-48  & 48$\times$48$\times$32             \\
																					& RPL-64  & 64$\times$64$\times$32             \\
																					& RPL-96  & 94$\times$96$\times$32             \\ \midrule
																			Downstream & NCC    & 48$\times$48$\times$48             \\
																					& NCS    & 64$\times$64$\times$32             \\
																					& LCS    & [512$\times$512$\times$(74$\sim$987)]$\ast$0.5 \\ \bottomrule
	\end{tabular}}
	\label{tab:input_size}
	\end{table}

\begin{figure*}[!ht]
	\centering
	\includegraphics[width=0.75\linewidth]{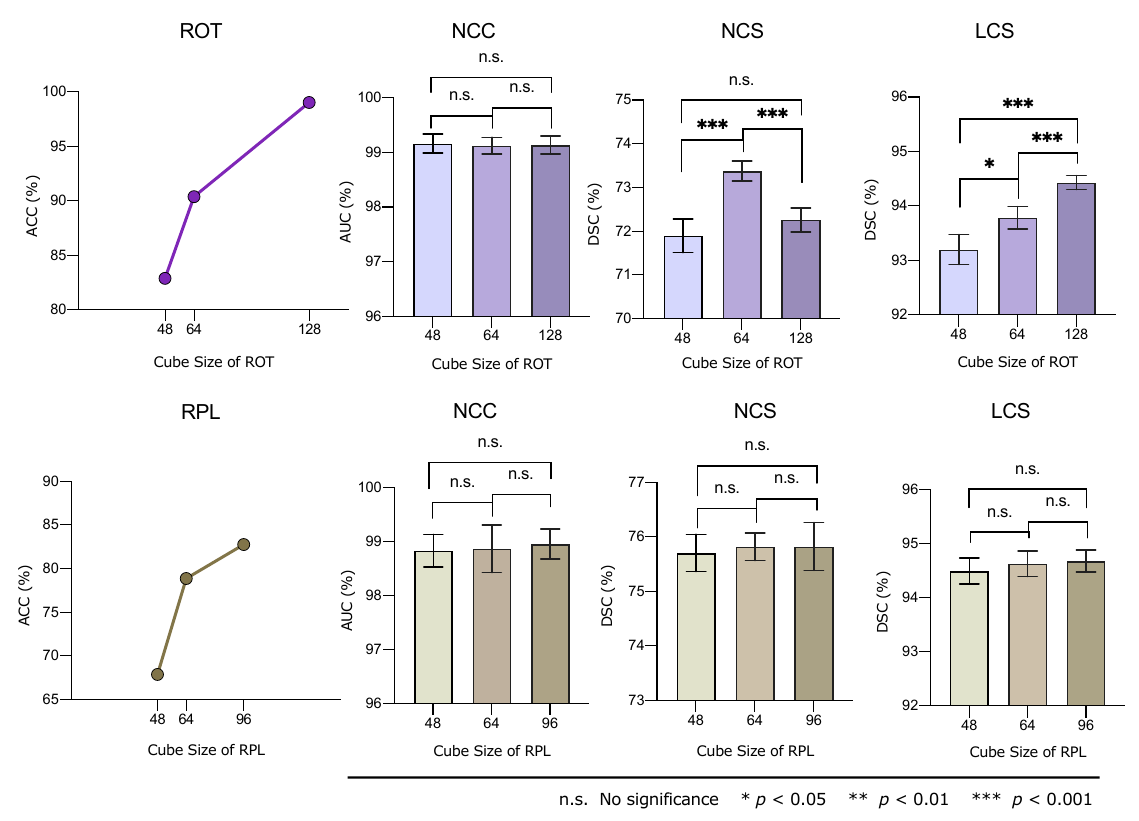}
	\caption{The impact of input size for predictive SSL methods. For pretext tasks, the classification accuracy with regard to varying input sizes is evaluated on the validation set. For target tasks, we evaluate the AUC score and DSC for classification and segmentation tasks respectively on each test set.}
	\label{fig:input_size}
\end{figure*}

In a predictive pretext task, the size of input volumes determines the amount of spatial information from which the network extracts features, thus influencing the expressive power of SSL features. As a crucial hyperparameter for SSL pretraining, the input size adopted in pretext tasks needs to be thoroughly investigated. However, related works \citep{taleb20203d, zhuang2019self} only supply a brief description of the choice of input size. To assess the impact of input size on SSL features, we pretrain SSL models using inputs with different sizes (see Table~\ref{tab:input_size}) and examine them in various target tasks (see Fig.~\ref{fig:input_size}). Below are our key findings:

\textbf{The accuracy of proxy tasks increases as the input size grows.} The logic is that it will be easier for networks to predict transformations with more available spatial information. 

\textbf{The accuracy of proxy tasks and the performance of downstream tasks are not simply positively correlated.} Specifically, higher accuracy of proxy tasks does not imply better target performance. When the input size is reduced to 48, we suspect that valid SSL features could not be well-learned by the ROT model (At least the RPL model takes two patches as input instead of a single patch). This can explain why the ROT model with small inputs performs poorly in both NCS and LCS.

\textbf{The sensitivity of input size varies for different SSL methods and target tasks.} For the classification task, the variation caused by varying input sizes is negligible for both ROT and RPL. However, segmentation tasks tend to exhibit distinctive performance when using ROT. That is, the ROT pretrained model is very sensitive to the input size of upstream data in segmentation, and it performs best when the input size of upstream data is closest to the downstream data. In contrast, the RPL pretrained model is robust to the input size.

To prevent the risk of performance degradation due to transferable features being sensitive to inconsistent input size, a simple proposal is to keep the input sizes of pretraining and fine-tuning data as close as possible.

\subsection{Feature analysis of SSL models}

Convolutional neural networks are generally assumed to extract hierarchical features, where the low/middle layers contain general features and the higher layers extract more task-specific features \citep{zeiler2014visualizing}. Based on the above understanding, the benefits of transfer learning have been proved to stem from pretrained low/mid-level feature reuse in target tasks \citep{neyshabur2020being, zhao2020makes}. Major applications of SSL methods are still obeying the transfer learning paradigm at present, so the same conclusion holds for SSL pretraining \citep{asano2019critical,islam2021broad}. To access a better grasp of the impact of different SSL pretrained features on medical imaging tasks, we carry out the following quantitative and qualitative analyses.

\begin{figure*}[!ht]
    \centering
    \includegraphics[width=0.85\linewidth]{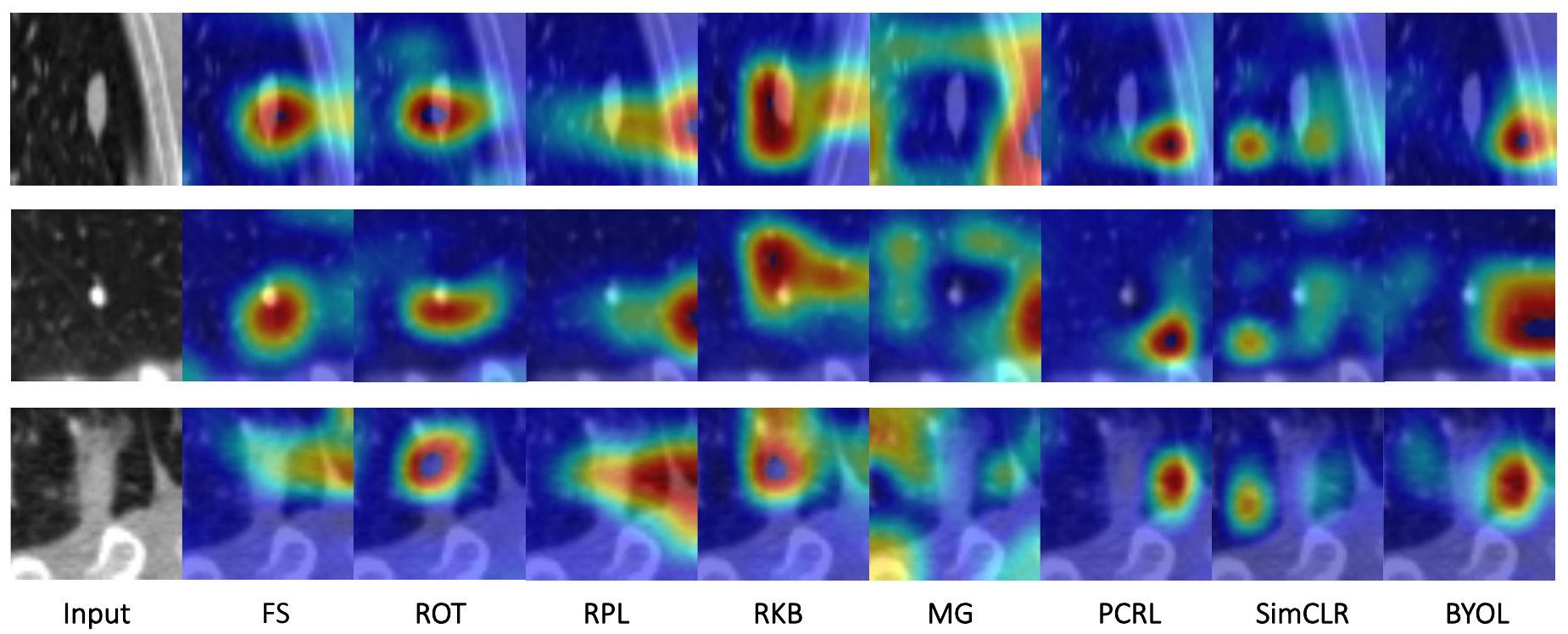}
    \caption{Grad-CAMs of different SSL methods in NCC. The left column exhibits original images containing lung nodules.}
    \label{fig:CAM}
\end{figure*}

\textbf{A. Feature visualization analysis.}
To investigate the difference between reused features of various SSL methods for the classification task, we visualize focal regions associated with the network decisions via Gradient-weighted Class Activation Maps (Grad-CAMs). The results of networks trained on NCC dataset with different SSL approaches are shown in Fig.~\ref{fig:CAM}. Our observations are as follows:
\begin{itemize}
	\item Predictive SSL models (ROT, RPL, RKB) look exactly at nodules. Compared to the constrictive activation region in ROT, RPL and RKB models pay additional attention to the region around the nodule. The reason might be that RPL and RKB involve predicting relative positions and therefore demand more context information while ROT model focuses more on the discriminative structure of objects.
	\item The focus of generative SSL models is prone to be on backgrounds rather than foreground objects. This is reasonable because the reconstruction task is bound to focus on subtle details. Consequently, the generative SSL leads to better performance gains in segmentation tasks rather than classification tasks, as the fine-grained features are more desired for delineating objects.
    \item Although contrastive SSL models have noticed partial nodules, most of the attention is distracted to surrounding regions. Such off-focus is caused by the contrastive learning mechanism that depends on maximizing mutual information between positive pairs, whereas a rigorous interpretation remains a mystery.
\end{itemize}
\begin{table*}[!t]
	\renewcommand{\arraystretch}{1}
	\caption{Comparison of feature-reuse rates of different SSL methods on two downstream datasets. For each method, we report the CKA similarity scores between the pretrained and fine-tuned models in terms of 4 layers. In each column, the CKA scores above 70\% are highlighted in bold. The number after ``ROT" and ``RPL" stands for the input size in the axial view.}
	\centering
    \resizebox{0.8\textwidth}{!}{
	\begin{tabular}{cccccccccc}
	\toprule
	\multirow{2}{*}{Pretraining}    & \multirow{2}{*}{Method} & \multicolumn{4}{c}{NCC}                                  & \multicolumn{4}{c}{NCS}                                  \\ \cmidrule(r){3-6} \cmidrule(r){7-10}  
									 &                         & Layer1         & Layer2         & Layer3         & Layer4         & Layer1         & Layer2         & Layer3         & Layer4         \\ \toprule
	\multirow{1}{*}{Predictive SSL}  & ROT-48                   & \textbf{0.954}          & \textbf{0.747}         & 0.493          & 0.292          & 0.878          & 0.867          & 0.758          & 0.705          \\
									 & ROT-64                   & \textbf{0.944}          & \textbf{0.815}         & \textbf{0.599}          & 0.243          & \textbf{0.945}          & 0.901          & 0.872          & 0.745          \\
									 & ROT-128                  & 0.688          & 0.646          & 0.412          & \textbf{0.611}         & 0.805          & 0.838          & 0.833          & 0.730          \\
									 & RPL-48                   & 0.805          & \textbf{0.692}          & 0.520          & 0.174          & \textbf{0.979} & \textbf{0.970} & \textbf{0.956}          & 0.786          \\
									 & RPL-64                   & 0.815          & 0.648          & 0.541          & 0.197          & \textbf{0.975}         & \textbf{0.954}          & \textbf{0.952}          & \textbf{0.902}          \\
									 & RPL-96                   & \textbf{0.823}          & 0.628          & 0.489          & 0.280          & \textbf{0.971}         & \textbf{0.955}          & \textbf{0.965} & \textbf{0.938} \\
									 & RKB                     & \textbf{0.968} & \textbf{0.863} & \textbf{0.706} & \textbf{0.625} & 0.897          & 0.905          & 0.893          & 0.788          \\ \midrule
	\multirow{1}{*}{Generative SSL}  & MG                      & 0.810          & 0.667          & 0.506          & 0.140          & 0.949          & 0.912          & 0.819          & 0.665          \\
									 & MG(w/o skip)            & 0.676          & 0.680          & \textbf{0.636}          & \textbf{0.415}         & 0.891          & \textbf{0.937}          & \textbf{0.929}          & \textbf{0.826}          \\
									 & PCRL                    & 0.728          & 0.461          & 0.171          & 0.113          & 0.937          & 0.863          & 0.757          & 0.643          \\
									 & PCRL(w/o skip)       & 0.652          & 0.483          & 0.320          & 0.172          & 0.865          & 0.888          & 0.723          & 0.626          \\ \midrule
	\multirow{1}{*}{Contrastive SSL} & SimCLR                  & 0.598          & 0.652          & 0.532          & 0.383          & 0.837          & 0.868          & 0.862          & 0.787          \\
									 & BYOL                    & 0.683          & 0.665          & \textbf{0.690}         & \textbf{0.463}          & 0.882          & 0.889          & 0.905          & \textbf{0.837}          \\ \bottomrule
	\end{tabular}}
	\label{tab:CKA}
	\end{table*}

\textbf{B. Feature similarity analysis.}
Intuitively, useful pretrained features for downstream tasks would be retained during the fine-tuning process. Thus, the extent of feature reuse can indicate the efficiency of an SSL algorithm in downstream tasks. Following the prior works \citep{neyshabur2020being, taher2022caid}, we use the Centered Kernel Alignment (CKA) \citep{kornblith2019similarity} to measure the feature similarity of the pretrained features and fine-tuned features on downstream tasks. To avoid interference caused by the disparity between upstream and downstream data, we examine the CKA score on NCS and NCC datasets, which are identical to the pretraining data. We extract features from conv1 to conv4 in the encoder of U-Net, denoted as Layers 1 to 4, and then pass them to a global average pooling layer to compute feature similarity. The average CKA scores are reported in Table~\ref{tab:CKA}. In particular, we add generative SSL models without skip connections in U-Net and predictive SSL models with varying input sizes to further investigate the idea in \ref{sec:input_size} and \ref{sec:u_shape_arch}. We first analyze the trends of diverse SSL methods in general, and then scrutinize the influence of input size and model structure:
\begin{itemize}
	\item For NCC, it is obvious that the lower features in the first three layers are more reusable in predictive SSL than generative and contrastive SSL. The middle-feature similarity in Layer4 is high in ROT, RKB and BYOL. In fact, it is precisely these three methods that yield the best AUC in NCC according to Table~\ref{tab:comparison}. As for NCS, RPL shows a remarkably high feature reuse score at each layer, accounting for the superior performance in Table~\ref{tab:comparison}. Besides, BYOL shows higher feature similarity in Layer4, accounting for its best results in NCS. By associating feature reuse with target performance, we declare that higher feature reuse is more productive for transfer learning.
	\item Overall, the feature-similarity variation due to input size is evident in ROT but negligible in RPL across two downstream tasks. This can further explain why the downstream performance is sensitive to ROT in Fig.~\ref{fig:input_size}. One exception is that the performance of NCC is robust to ROT. This may be because the significant improvement in the feature reuse score at Layer4 of the ROT-128 model compensates for the poor feature reuse at the first three layers.
    \item Regarding the U-shape structure, we discover that removing the skip connections in MG significantly increases the feature-reuse rate in higher layers, even though it may compromise a bit feature-reuse rate in lower layers. This observation solidly confirms our view in \ref{sec:u_shape_arch}, i.e. removing the skip connections suppresses the overfitting issue in a pure reconstruction task. However, for PCRL composed of multiple pretext tasks, removing skip connections would instead make it more difficult to learn valid transferable features, resulting in less feature reuse.

\end{itemize}

\section{Common training policies}
\label{sec:augmentation}

\begin{figure*}[!ht]
    \centering
    \includegraphics[width=0.7\linewidth]{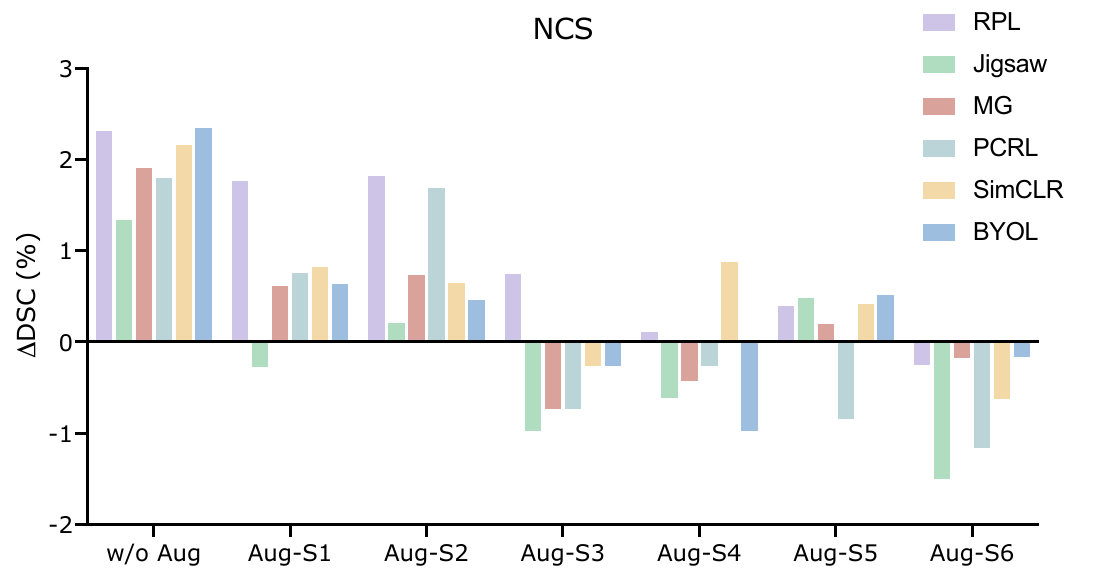}
    \caption{The effects of data augmentation on different SSL methods. $\Delta$DSC refers to the difference in DSC between an SSL method and the baseline. The data augmentation degree increases from ``Aug-S1" to ``Aug-S6".}
    \label{fig:aug}
\end{figure*}

Both previous experiments in natural image classification \citep{liu2021imbalance,yang2020rethinking} and our experiments in Section~\ref{sec:imbalance} provide some tentative evidence that one of the foremost benefits of SSL pretraining is the improvement in class imbalance. 
Regarding SSL as a strategy to tackle class imbalance, we wonder if it could be overlaid with other widely used training policies, including data resampling and augmentations, to achieve better performance. To this end, we evaluate the effects of stacking SSL with other strategies in NCC and NCS.
\changed{M1.8.4}{
\link{R1.8}}{Main takeaways in this section are summarized below: \begin{enumerate}
\item The combination of resampling and SSL techniques is preferred in severe class imbalance and low-data regimes, whereas solely resampling is preferred in slight class imbalance.~(Table~\ref{tab:resampling})
\item Strong augmentation diminishes the value of SSL pretraining.~(Fig.~\ref{fig:aug})
\end{enumerate}}

\begin{table*}[!t]
	\renewcommand{\arraystretch}{1}
	\caption{The performance gains of different training strategies over the baseline (trained from scratch) in NCC. The additional gain of SSL pretraining over mere resampling ($\textgreater$ 1\%) is highlighted in blue.}
	\setlength\tabcolsep{1.8pt}
	\centering
 \resizebox{0.8\textwidth}{!}{
	\begin{tabular}{cccccccccccc}
	\toprule
	\multirow{2}{*}{Imbalance Ratio} & \multirow{2}{*}{$N$} & \multicolumn{2}{c}{Mean-Resampling}     & \multicolumn{4}{r}{+MG}             & \multicolumn{4}{r}{+PCRL}           \\ \cmidrule(r){3-4} \cmidrule(r){5-8}  \cmidrule(r){9-12}  
										&                    & $\Delta_{AUC}$   & $\Delta_{Recall}$ & \multicolumn{2}{c}{$\Delta_{AUC}$} & \multicolumn{2}{c}{$\Delta_{Recall}$} & \multicolumn{2}{c}{$\Delta_{AUC}$}    & \multicolumn{2}{c}{$\Delta_{Recall}$} \\ \toprule
	\multirow{3}{*}{$r$=0.01}          & 5k                 & -4.21 & 19.10                       & \textcolor{blue}{(+6.10)} & 1.89  & \textcolor{blue}{(+13.28)}  & 32.38                    & \textcolor{blue}{(+2.77)}  & -1.44 & \textcolor{blue}{(+24.24)}  & 43.34             \\
										& 10k                & 9.96  & 37.00                      & \textcolor{blue}{(+12.57)} & 22.53 & \textcolor{blue}{(+17.31)} & 54.31                  & \textcolor{blue}{(+4.93)} & 14.89 & \textcolor{blue}{(+22.53)} & 59.53                    \\
										& 15k                & 17.21 & 52.66                      & \textcolor{blue}{(+1.96)} & 19.17 & \textcolor{blue}{(+17.75)} & 70.41                  & \textcolor{blue}{(+4.01)} & 21.22 & & 47.52                      \\ \hline
	\multirow{3}{*}{$r$=0.05}          & 5k                 & 4.14  & 45.16                      & \textcolor{blue}{(+10.20)}  &14.34 & \textcolor{blue}{(+3.93)}  &49.09                & \textcolor{blue}{(+2.33)}  & 6.47 &  & 38.38                      \\
										& 10k                & 22.94 & 47.25                       & & 22.98  &  & 47.01                     &  & 21.39   & & 41.78                      \\
										& 15k                & 21.05 & 44.13                      & & 21.25  & \textcolor{blue}{(+8.09)} & 52.22               &     & 19.04  & & 44.91                      \\ \hline
	\multirow{3}{*}{$r$=0.1}           & 5k                 & -6.20 & 40.21                      & \textcolor{blue}{(+4.68)} & -1.52  & & 34.99                      & \textcolor{blue}{(+2.94)} & -3.26 & & 26.37                      \\
										& 10k               & 16.42  & 47.78                       & & 16.12   & &  47.51                   &    & 16.45  & & 43.08               
										       \\
										& 15k                & 11.26 & 47.51                     &  & 11.53  & & 46.21                     &  & 9.98   &\textcolor{blue}{(+2.10)} & 49.61                      \\ \bottomrule
	\end{tabular}}
	\label{tab:resampling}
	\end{table*}

\subsection{The additive effect of SSL and resampling.}

As a simple yet effective strategy for class-imbalanced learning, data resampling over-samples data from rare classes or under-samples data from frequent classes to alleviate class bias \citep{cao2019learning}. Following \cite{he2021rethinking}, we adopt the mean resampling schedule for NCC, i.e. assigning an equal probability to sample each class during training epochs. We consider different class-imbalance settings as in Section~\ref{sec:imbalance} by changing the imbalance ratio $r$ and the number of training samples $N$. In each setting, we first train a network from scratch without any extra strategies as the baseline. Then, we use the mean resampling and SSL pretraining to retrain the network.

Table~\ref{tab:resampling} summarizes the performance gains from data resampling and SSL over the baseline in NCC. The AUC and recall scores focus on overall classification and positive class performance respectively. Note that the positive class is the minority class in NCC. An obvious downside of the mean resampling schedule is that there might be extremely few samples in the positive class (small $r$ or $N$). This means that resampling discards a large amount of the data to ensure equal sampling possibility in both classes, which can lead to overfitting the small remaining portion of the data. With an increasing number of positive classes, the mean resampling schedule tends to exhibit larger gains over the baseline. Nonetheless, these gains are limited since the performance seems to saturate when $r$ and $N$ are large enough, namely slight class imbalance. Based on the resampling, we further explore the impact of SSL pretraining. 

\textbf{The combination of resampling and SSL techniques is preferred in severe class imbalance and low-data regimes while solely resampling is preferred in slight class-imbalanced regimes.} As opposed to the poor performance of SSL pretraining under extreme class imbalance (small $r$) in Fig.~\ref{fig:figure5}, we observe that SSL pretraining with the resampling schedule can boost performance significantly. This indicates a complementary effect of SSL and resampling on severe data-imbalanced learning: SSL can improve the drawback of overfitting a few samples in mean resampling, while mean resampling can balance the skewed SSL representation distributions. In scenes with minor data imbalance, however, SSL methods generally have little impact on the overall performance. Actually, using only SSL pretraining lags behind the mean resampling schedule a lot.

\subsection{The additive effect of SSL and data augmentation.}

Data augmentation plays an essential role in training CNNs, involving preventing overfitting and enhancing generalization performance. Additionally, it can mitigate the data imbalance issue by reducing overfitting to the observed samples, most of which belong to the majority classes. We employ six compositions of data augmentation at different degrees in NCS and investigate the stacking effects. Table~\ref{tab:resampling} plots the performance gains from SSL pretraining with different data augmentation policies over the baseline in NCS.

\textbf{Strong augmentation diminishes the value of SSL pretraining.} We find that SSL only provides clear improvements without any data augmentation or with weak data augmentation. But as we increase the strength of data augmentation, the benefit of SSL diminishes. Several SSL techniques even hurt model performance when strong data augmentations are applied. We hypothesize that strong augmentation already advances class-imbalanced learning which might cancels out the gains introduced by SSL.

\section{Discussions and Future Directions}\label{sec:discussion}

\subsection{The effectiveness of SSL for medical image analysis}
Our results in Table~\ref{tab:comparison} suggest that contrastive learning tasks exhibit more promising performance, albeit no SSL task always performs best across all the downstream tasks. In our experiments, we directly apply prevalent contrastive learning frameworks in natural scenes to medical imaging tasks. Since multiple objects always exist in one medical image, we ensure that two crops in a positive pair overlap each other. Surprisingly, such models exceed most of the models pretrained with carefully-designed tasks across all three tasks. \changed{M2.5.2}{\link{R2.5}}{However, the main limitation of these simply extended contrastive learning methods is that they tend to focus more on global features instead of fine-grained information. As mentioned in Section~\ref{sec:ours_review}, some re-designed contrastive learning approaches with local-attention mechanisms have been proved to achieve much better performance for dense prediction tasks than the original global-attention contrastive learning counterparts.} 

For other SSL types, predictive and generative tasks generally show distinct performance in different target tasks. From the perspective of task design principles, there exists one intrinsic defect in current predictive and generative SSL tasks: the objective of many pretext tasks is so harsh that even a medical expert cannot accomplish it, which departs the purpose of self-discovering semantically useful information contained in medical images. Two SSL tasks are typical for our discussion: Jigsaw and MG. In Jigsaw, 3D volumetric data is always split into dozens of cubes and then shuffled in one of the hundreds of orders. 
There is a high probability that a cube contains only a small branch of an organ. Additionally, some small misaligned cubes do not affect the main anatomy at all. Hence, it is unjustified to force the network put dozens of cubes in exactly the right order. In MG, the network is tasked with recovering every pixel in the medical data even if some patch is meaningless for clinical diagnosis. There are many possible appearances for masked areas in line with the correct anatomical structure rather than only the original ground truth. Due to such rigid self-labelling, the network tends to find shortcuts or trivial solutions, failing to mine valuable information. Therefore, we believe it is necessary to develop SSL tasks to focus on salient prior knowledge, such as constructing surrogate labels with some variability but still obeying the pathological prior.

Recently, there is a growing interest in integrating different SSL tasks into a framework to achieve better performance \citep{dong2021self, taher2022caid}, e.g. combining contrastive and generative tasks to learn both global and local features. It is acknowledged that multi-task learning prevents representations from being skewed towards a particular pretext task and pushes the representations to be more robust and high-quality. However, such paradigms render the self-training stage complicated and costly. Besides, the weight of each sub-task is not trivial to search \citep{jin2021automated}. We expect to see more simple yet effective work for medical imaging in the future.

\subsection{The objective value of SSL for medical applications}

Compared with natural images, medical images have two characteristics that challenge the performance and training efficiency of deep learning: (1) small data scale caused by expensive data collection and annotation costs. (2) data bias arising from minority disease cases in the clinical routine inspection or small lesion areas in the whole imaging region; Most of the existing works have demonstrated the advantages of SSL in mitigating the issue (1), whereas fewer works have systematically assessed the implications of SSL on the issue (2). Based on our extensive experiments, we rethink the value of SSL in both aspects:

\begin{itemize}

\item{There is now broad consensus that SSL yields better performance in downstream tasks with very limited labelled data (maybe only a few dozen samples), but the returns are diminishing as the downstream labelled training set grows in size \citep{tajbakhsh2019surrogate, taleb20203d}. 
As the community of medical vision expands, medical image datasets of larger size have been or are being created. To achieve remarkable results on moderate-sized annotated medical datasets, it is still a long way to go for SSL.}

\item{Our theoretical model in Section~\ref{sec:imbalance} shows that both the class imbalance in labelled data and the potential data imbalance in unlabelled data used for SSL pretraining affect the estimation of the target classifier. Although the modelling of deep networks is much more complex than our assumed theoretical model, we can get a clear intuition of how SSL behaves under data imbalance. Our experiment results confirm that SSL significantly boosts the performance of rare classes but yields a marginal impact on frequent classes in both classification and segmentation tasks, thereby improving class-imbalanced learning. This phenomenon seems to reveal the essence of SSL, which is to prevent the network from overfitting frequent samples. It is natural to ask if we could leverage other commonly used strategies to improve the overfitting issue, including data resampling and data augmentation. For that purpose, we investigate the additive effect of SSL with mean resampling and augmentation policies, respectively. The former results indicate that the performance of SSL can be compounded by mean resampling under extreme data imbalance. However, SSL pretraining is worthless and the simple mean-resampling performs well when the data imbalance is not severe. The latter results suggest that the value of SSL loses when using stronger data augmentation. This enlightens researchers to regard SSL pretraining as a tool to avoid a network overfitting frequent classes. However, it remains a mystery why some SSL methods achieve poorer performance than the baseline when using strong augmentation. This phenomenon is consistent with a previous finding in \cite{zoph2020rethinking} that supervised pretraining also hurts the downstream performance when using strong augmentation. We believe this is not a coincidence, and the interplay between pretraining and data augmentation deserves deeper analysis.}

\end{itemize}

\subsection{Towards better transfer schemes from SSL pretrained models to downstream medical tasks}

The success of SSL pretraining depends on not only an effective representation learning via self-supervision signals but also how well the pretrained weights are reused in downstream learning. Looking back, we notice that the transfer process was often neglected in the previous literature. To find better transfer schemes, we discussed two relevant aspects in Section~\ref{sec:u_shape_arch}. First, we explore the role of each module in the pretrained model with a U-shape architecture, including the encoder, decoder, skip connections and BN layers. The results show that the pretrained encoder is exactly what needs to be transferred. On the contrary, the pretrained decoder is dispensable, which contradicts claims from some SSL works that the pretrained decoder is superior to a randomly-initialized decoder \citep{tao2020revisiting}. Given the lack of experimental verification in these works, our finding is more convincing. Besides, we prove that removing the skip connections in generative SSL leads to better representations in the encoder. As for the BN layers, we recommend to re-collect BN statistics for target data at the test stage since the domain gap between the source data and the target data might degrade performance. Finally, we compare different fine-tuning schemes used in previous works. It turns out that full fine-tuning with a small learning rate is the best choice currently.

Despite our recommendations for professionals on better implementations of transfer, there is a large room for future researchers to design better weight reuse schemes. One potentially useful direction is the development of adaptive optimizers that can adjust the learning rate of different parameters according to their behaviours in downstream training, ensuring that expressive pretrained representations are retained and under-learned representations are updated. Another research direction is to design multi-channel BN layers oriented to different learning stages, following some domain adaptation techniques \citep{chang2019domain, zhou2022generalizable}. Overall, we believe there is an urgent need for research teams to improve current transfer learning paradigms.

\section{Conclusion}
\label{sec:conclusion}
We systematically examine the value of SSL in medical image analysis and present practical guidelines for SSL professionals throughout the entire pretraining and fine-tuning process. On the positive side, we have demonstrated the impressive achievements of SSL in advancing class-imbalanced learning. This is of great clinical value because the diagnosis rate of true positive disease cases is crucial. However, on the negative side, we argue that SSL pretraining is not useful always: (1) the benefits of SSL can be superseded by other frequently-used strategies in some applications, such as data resampling and augmentation. (2) the performance gains from SSL are very limited in middle-size downstream datasets, compared to the insufficient annotated datasets in previous studies. We hope that our in-depth study based on around 250 experiments running over 2000 GPU hours will help to access better capabilities of SSL and shed light on promising research directions for the community to move forward.

\section*{Declaration of Competing Interest}
The authors declare that they have no known competing financial interests or personal relationships that could be appeared to influence the work reported in this paper.

\section*{CRediT authorship contribution statement}
\noindent
\textbf{C.Z:} Conceptualization, Methodology, Software, Validation, Writing- original draft; \textbf{H.Z:} Supervision, Conceptualization; \textbf{Y.G:} Supervision, Conceptualization, Writing - review and editing.

\section*{Acknowledgement}
This work is supported in part by the Open Funding of Zhejiang Laboratory under Grant 2021KH0AB03, in part by the Shanghai Sailing Program under Grant 20YF1420800, and in part by NSFC under Grant 62003208, and in part by Shanghai Municipal of Science and Technology Project, under Grant 20JC1419500 and Grant 20DZ2220400.

\appendix
\section{Implementation details for all tasks}
\label{sec:appen_implementation}

Our attached SSL code base allows training and evaluation for various pretext and downstream tasks in a unified framework. It is easy for researchers to implement new tasks on our code base, thereby making it convenient to compare their proposed methods with existing SSL methods. Our implementations rely on Pytorch \citep{paszke2019pytorch} and some APIs for medical imaging. Details can be found in the README.md file. 

\subsection{Pretext tasks}

\textbf{3D pretraining data.}
Following \cite{zhou2021models}, we pretrain all the 3D SSL models on the LUNA 2016 dataset \citep{LUNA16} which is composed of 10 subsets with 888 low-dose chest CT scans. 
The 1-7 subsets (623 images) are used for pretraining and fine-tuning, with 1-5 subsets (445 images) as the training set and 6-7 subsets (178 images) as the validation set. The remaining 8-10 subsets (265 images) constitute the test set for target tasks. For each CT scan, we apply an intensity window at [${}-$1000, 1000] and then normalize it to [0, 1]. 

\textbf{2D pretraining data.}
We select 28k high-quality fundus images from EyePACS dataset \citep{eyepacs} based on the gradability scores in \cite{voets2019reproduction}. We resize each 2D image to a unified resolution of 512 $\times$ 512 and normalize it to [0, 1]. Note that the target task EPC also utilize the images from EyePACS dataset. To avoid test data leakage, we keep the test images unseen for pretraining.

\textbf{Predictive SSL.}
The predictive SSL models are trained by Adam optimizer \citep{kingma2014adam}. The learning rate is initially set as 1e-3 and then decayed by a factor of 10 at 250 epochs. To prevent overfitting, the training stops if no improvement is observed in the validation set over certain epochs. For the 3D input processing, we center-crop each CT volume to a unified size of [320, 320, 74] ([H, W, D]). This step ensures that: (a) the images of the same size allow batch training; (b)most of the empty areas were removed according to the CT image prior. For each 2D image, we perform center-cropping with the size of 360 $\times$ 360 to eliminate the edges of the field of view, thereby avoiding shortcut solutions.
\begin{itemize}
	\item \textbf{RPL:} We split 3D volumes into $3 \times 3 \times 3$ patches and 2D images into $3 \times 3$ patches. 
 To prevent shortcut solutions such as edge continuity, we leave a random gap of 3 pixels per axis between adjacent patches.
	\item \textbf{ROT:} 
    We randomly crop small patches from each CT volume as the inputs.
	\item \textbf{Jigsaw, RKB, RKB+:} We split 3D volumes into $2 \times 2 \times 2$ patches and 2D images into $3 \times 3$ patches. Then, we select 100 permutations with the largest Hamming distance from all the $8!$ permutations for the jigsaw puzzle task.
\end{itemize}

\textbf{Generative SSL.}
We train generative SSL models (AE, MG) by SGD optimizer \citep{zhang2004solving} with an initial learning rate of 1e0. We employ a learning rate annealing strategy namely ReduceLROnPlateau. In specific, the learning rate is decayed by half when no improvement is observed in the validation set over a specified number of epochs. Especially for 3D scans, we randomly crop patches with the size of [64, 64, 32] and exclude those that are empty (air) or contain full tissues according to the intensity values. Finally, the training set consists of 14159 patches with the size of [64, 64, 32] and the validation set comprises 5640 patches with the size of [64, 64, 32].

\textbf{Contrastive SSL}
We train contrastive SSL models (PCRL, SimLCR, BYOL) by SGD optimizer \citep{zhang2004solving} with an initial learning rate of 1e-3. The cosine annealing strategy is adopted to schedule the learning rate. In typical contrastive learning for object-centric natural images, random cropping and some other augmentations are applied to one image to generate a positive pair. However, such pair generation might lead to model confusion for medical images containing multiple objects. For effective contrastive learning in CT images, a crucial trick is to crop positive pairs with overlaps. Specifically, we randomly crop a pair of patches with the size of [64 64, 32] in one volume and compute their IoU (Intersection over Union) score. Only pairs with the IoU score larger than 0.25 can be kept. For each training volume, such operations are repeated several times. Finally, the training dataset is composed of 9968 positive pairs.

\subsection{Target tasks}
We utilize three 3D medical imaging tasks that are the most frequent in existing publications of SSL: NCC, NCS and LCS. Besides, we also use the MSD-liver dataset to explore multi-class segmentation. Note that we keep the dataset partitioning protocols the same as \citet{zhou2021models}. Moreover, we implement two retinal fundus imaging tasks in the benchmarking platform: EPC and DVS. The training details for each are as follows.

\textbf{NCC:}
The dataset for NCC is from LUNA 2016 \citep{LUNA16}, which contains 5,510,166 candidate locations for the lung nodule false positive reduction task. According to the location annotations, we can obtain candidate cubes with the size of [48, 48, 48] as inputs. The dataset splitting has been introduced at the beginning of this section, leading to a training set with 282568 cubes, a validation set with 109482 cubes and a test set with 166225 cubes. The false positive cubes are labelled as class 0 while the true positive cubes are tagged as class 1. The number of training samples for class 0 and class 1 is 274637 and 7931, respectively. Obviously, NCC is a severely imbalanced binary classification task. Thus, we perform a mean resampler to ensure the equal number of class 0 and class 1 at each epoch. During training, we adopt the binary cross entropy loss to learn classifiers. Adam optimizer \citep{kingma2014adam} with a learning rate of 1e-3 is used for optimization. 

\textbf{NCS:}
The dataset for NCS is from the Lung Image Database Consortium image collection (LIDC-IDRI) \citep{LIDC}, which consists of 1018 lung CT scans. For each scan, the nodules are labelled with binary masks. We split these scans into training (510), validation (100), and test (408) sets. After re-sampling the volumes to 1-1-1 spacing, each nodule is extracted by a $64 \times 64 \times 32$ crop. We adopt the dice loss function to train the model. Adam optimizer \citep{kingma2014adam} with a learning rate of 1e-3 is used for optimization. The training stops if no improvement has been observed in the validation set for dozens of epochs.

\textbf{LCS:}
The dataset for LCS is from MICCAI 2017 LiTS Challenge \citep{LiTS}, which comprises 130 abdominal CT scans. The preprocessing stage includes clipping with a HU window of [-200, 200], normalization to [0, 1] and resampling to 1-1-1 spacing. Due to the limited GPU memory, a 0.5x downsampling of the axial view is performed on each image. We split all scans into training (100), validation (15) and test (15) sets. A combination of binary cross entropy loss and dice loss is adopted to train the model. Adam optimizer with an initial learning rate of 1e-2 is used for optimization. At the 50th and 150th epoch, the learning rate is reduced by a factor of 10. The training stops if no improvement has been observed in the validation set for dozens of epochs.

\textbf{MSD-Liver:}
The dataset for MSD-Liver is from the Medical Segmentation Decathlon Challenge (MSD) \citep{MSD}, which consists of 130 labelled abdominal CT scans. The images are the same as the LCS but here we use both liver and tumor annotations, which makes MSD-liver task more challenging than LCS. We applied the data preprocessing scheme in nnUNet \citep{isensee2021nnu} to generate inputs. The data splitting is the same as LCS. We combine the cross entropy with dice loss functions to train the model. Adam optimizer with an initial learning rate of 1e-2 is used for optimization. At the 150th and 300th epoch, the learning rate is reduced by a factor of 10. The training stops if no improvement has been observed in the validation set for dozens of epochs.

\textbf{EPC:}
For EPC, we randomly sample 10\% of the entire EyePACS dataset \citep{eyepacs} (3511/1090/4265) for training/validation/testing. The preprocessing steps are the same as pretraining. To consider the rankings of different levels of diabetic retinopathy grading, we solve this task on a regression model (using a single neuron for the output) and adopt the mean square error (MSE) loss function. Adam optimizer with a learning rate of 1e-3 is used for optimization. The training stops if no improvement has been observed in the validation set for dozens of epochs.

\textbf{DVS:}
The DRIVE dataset \citep{drive} is split to 20/5/15 for training/validation/testing. Adam optimizer with the learning rate of 1e-3 is used for optimization. The images with size of 565 $\times$ 584 are cropped to 560 $\times$ 576 and normalized to the range [0, 1] before training.  
 We use the dice loss function and Adam optimizer with a learning rate of 1e-3 to train the model.
The training stops if no improvement has been observed in the validation set for dozens of epochs.

\section{Network architectures for all tasks}
We use the encoder of 3D U-Net \citep{cciccek20163d} for all 3D tasks. It includes four 3D convolutional blocks with 64, 128, 256 and 512 channels, respectively. For 2D tasks, we adopt the 2D U-Net \citep{ronneberger2015u} encoder consisting of five blocks with 64, 128, 256, 512 and 1024 channels. On the top of the encoder, we add specific network heads or decoders for different tasks, detailed below.

\textbf{ROT, NCC, EPC:}
The network head of ROT or NCC is a classifier constituted by a global average pooling layer and one hidden fully connected layer with 1024 (3D) / 2048 (2D) neurons.

\textbf{RPL:}
The network head of PRL consists of two branches sharing parameters but for reference patches and random patches respectively, and a classifier. First, a global average pooling layer is added after the encoder to get a dense hidden layer. Then, in two branches, a fully connected layer followed by a Batch Normalization layer get two $K_{RPL}$-dimensional representations separately from the reference patch and random patch. We set $K_{RPL}=1024$ for 3D RPL and $K_{RPL}=2048$ for 2D RPL. Next, these two vectors are concatenated and input to the classifier that is the same as the ROT.
\begin{table*}[!t]
\renewcommand{\thetable}{E.10} 
	\renewcommand{\arraystretch}{0.8}
	\caption{The effect of network architectures in RKB. The number after ``fc'' stands for the number of neurons. The ``Rot-ver'' and ``Rot-hor'' are short for vertically rotation task and horizontal rotation task respectively.}
	\centering
 \resizebox{0.8\textwidth}{!}{
	\begin{tabular}{ccccccc}
	\toprule
	 \multicolumn{3}{c}{Network heads \textcolor{blue}{(\# of params / M)}}                                            & \multicolumn{3}{c}{ACC (\%) on the validation set} \\ \cmidrule(r){1-3}\cmidrule(r){4-6}
									   Order classifier          & Rot-ver classifier      & Rot-hor classifier      & Order    & Rot-ver    & Rot-hor     \\ \midrule
	fc-1024-1024-100 \textcolor{blue}{(1.68)} & fc-1024-1024-8 \textcolor{blue}{(1.58)} & fc-1024-1024-8 \textcolor{blue}{(1.58)} & 35.52    & 49.09      & 51.29       \\ \midrule
fc-1024-1024-100 \textcolor{blue}{(1.68)}  & fc-1024-8 \textcolor{blue}{(0.53)}     & fc-1024-8 \textcolor{blue}{(0.53)}      & 94.94    & 98.31      & 96.28       \\ \bottomrule
	\end{tabular}}
	\label{tab:head_arch}
	\end{table*}

\textbf{Jigsaw, RKB, RKB+:}
The network heads of Jigsaw, RKB and RKB+ include multiple shared branches for partitioned cubes and classifiers for several sub-tasks. Firstly, a $K_{Jig}$-dimensional representation is extracted for each cube through a global average pooling and a fully connected layer followed by a Batch Normalization layer. We set $K_{Jig}=64$ for 3D RPL and $K_{Jig}=512$ for 2D RPL. For each image, the representations of all cubes are then concatenated and input to the order, rotation and mask classifiers. It is noteworthy that RKB and RKB+ are specifically designed for 3D volumetric data. Considering that the order task is more complex than the other two, the order classifier is composed of two hidden layers with 1024 neurons while the rotation and mask classifiers only contain one hidden layer with 1024 neurons.
 
\textbf{MG, AE, NCS, LCS, DVS:}
For reconstruction tasks or segmentation tasks, we use the standard 2D or 3D U-Net. We also implement U-Net versions without skip connections for reconstruction tasks. 

\textbf{SimCLR, BYOL:}
The network head of SimCLR and BYOL aims to project the input to the embedding space. For SimCLR, the feature projector consists of one hidden fully connected layer with 512 (3D) / 1024 (2D) neurons and a 256-dimensional output layer. BYOL requires two projectors and one predictor, where the structure of each is identical to the projector in SimCLR. 

\textbf{PCRL:}
The network head of PCRL contains three branches: a U-Net decoder to recover images, a projector with 128 hidden neurons in to extract representations for contrastive learning and a transformation module with 256 hidden neurons to generate the explicit controlling vector.
\begin{table}[!h]
 \renewcommand{\thetable}{C.9}
	\renewcommand{\arraystretch}{0.85}
	\caption{Notations for data augmentations used in NCS. Here, p denotes the probability for augmentation.}
	\centering
 \resizebox{0.48\textwidth}{!}{
	\begin{tabular}{c|l}
	\hline
	Augmentation & \multicolumn{1}{c}{Description}                                                                                                                                                                                   \bigstrut\\ \hline
	Aug-S1              & random flipping (p = 0.5)                                                                                                                                                                                         \bigstrut\\ \hline
	Aug-S2              & rotating by 90 degree (p = 0.5)                                                                                                                                                                                  \bigstrut\\ \hline
	Aug-S3              & \begin{tabular}[c]{@{}l@{}}random flipping (p = 0.5),\\ random rotating by degrees within {[}-10, 10{]} (p = 0.3)\end{tabular}                                                                                    \bigstrut\\ \hline
	Aug-S4              & \begin{tabular}[c]{@{}l@{}}rotating by 90 degree (p = 0.5),\\ random rotating by degrees within {[}-20, 20{]} (p = 0.3)\end{tabular}                                                                              \bigstrut\\ \hline
	Aug-S5              & \begin{tabular}[c]{@{}l@{}}random flipping (p = 0.5), \\ rotating by 90 degree (p = 0.5),\\ random rotating by degrees within {[}-30, 30{]} (p = 0.6)\end{tabular}                                               \bigstrut\\ \hline
	Aug-S6              & \begin{tabular}[c]{@{}l@{}}random flipping (p = 0.5), \\ rotating by 90 degree (p = 0.5),\\ random rotating by degrees within {[}-30, 30{]} (p = 0.6),\\ random scaling in {[}0.8, 1.2{]} (p = 0.4)\end{tabular} \bigstrut\\ \hline
	\end{tabular}}
	\label{tab:aug}
	\end{table}
\section{Augmentation schemes}
\textbf{NCC:}
On account of the very limited number of the true positive class, we augment the original true positive samples by 10 times with random rotating, shifting and flipping.

\textbf{NCS:}
When exploring the additive effect of data augmentation with SSL in NCS, we implement six data augmentation schemes from ``Aug-S1" to ``Aug-S6" (see Table~\ref{tab:aug}). Since random rotating by 90 degrees yields more different views than only flipping, we consider S2 stronger than S1. It is evident the strength of data augmentation increases from S1 to S6.

\textbf{Contrastive SSL:}
Many data augmentation strategies commonly used in natural images are inappropriate for gray-scale medical images, such as color jitter and grayscale. Thus, we only apply random flipping, rotating, small-range scaling and Gaussian blur on the positive pairs for pretraining on LUNA 2016 dataset.
 \section{Statistical significance analysis}
 \label{sec:t_test}
\begin{figure*}[!t]
   \renewcommand{\thefigure}{D.14}
    \centering
    \includegraphics[width=0.66\linewidth]{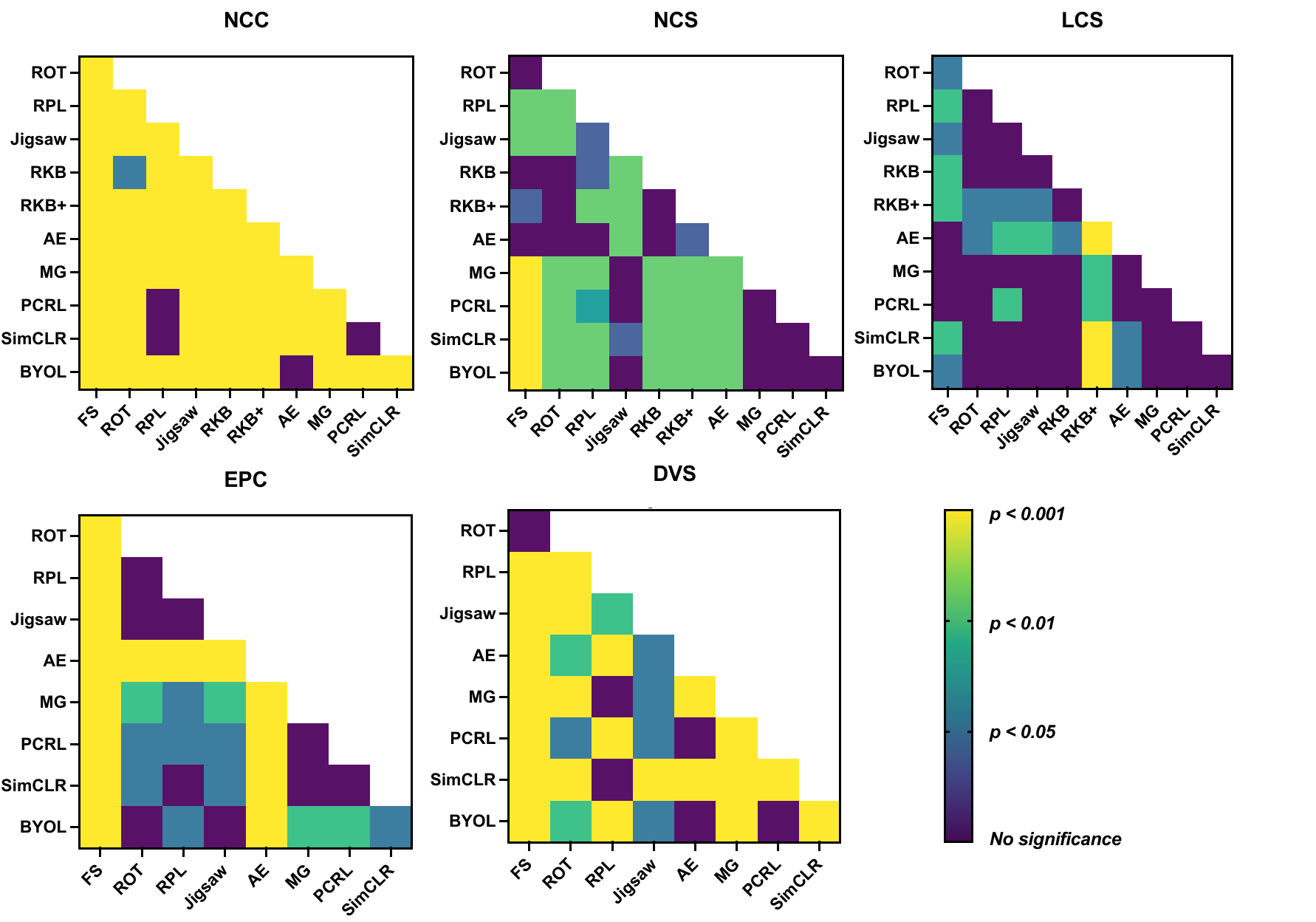}
    \caption{The paired t-test for all SSL methods in Table~\ref{tab:comparison}. Here, the evaluation metrics for NCC, NCS, LCS, EPC and DVS are ACC, DSC, DSC, Kappa and SEN, respectively.}
    \label{fig:exp_t_test}
\end{figure*}

 In Table~\ref{tab:comparison}, we have compared ten SSL methods across three datasets. Here, we perform paired t-test and report the $p$-values in Figure~\ref{fig:exp_t_test} to assess the statistical significance.
\begin{figure*}[!]
\renewcommand{\thefigure}{E.15}
\centering
\includegraphics[width=0.9\linewidth]{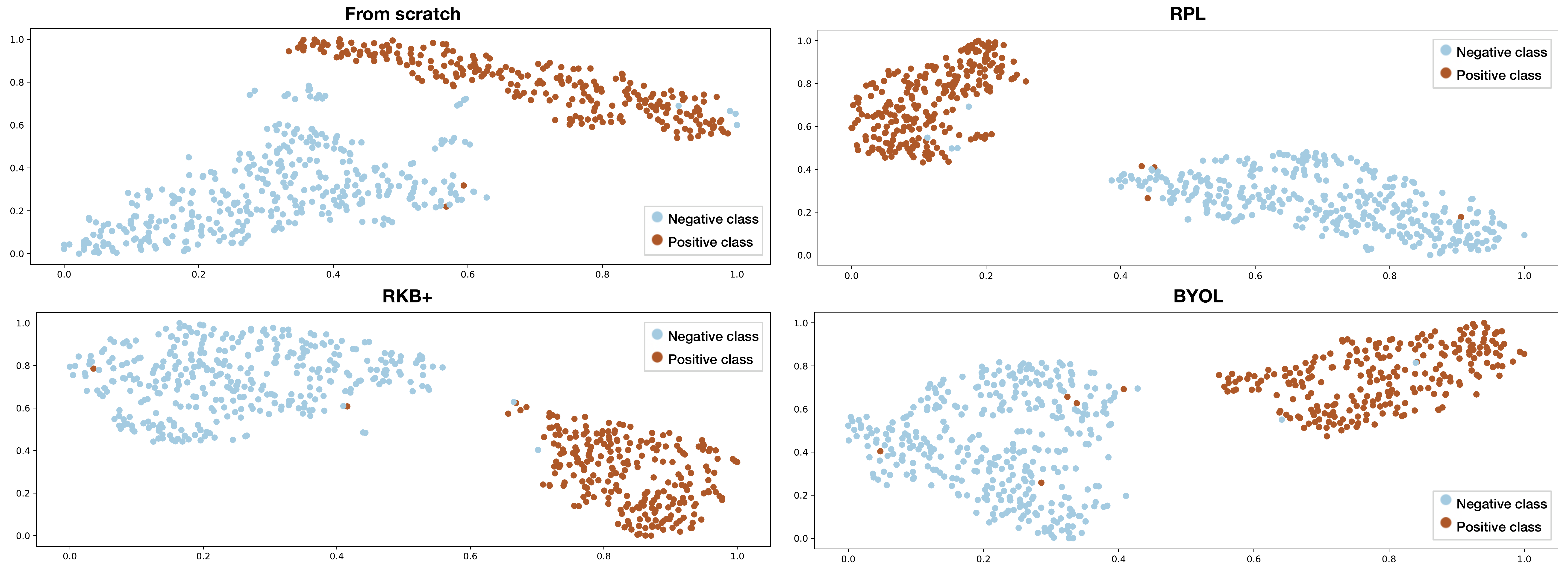}
\caption{t-SNE visualization of embeddings obtained by models pretrained with different SSL tasks in NCC.}
\label{fig:tsne_ncc_appen}
\end{figure*}
\section{Detailed experiments}
\begin{figure*}[!ht]
\renewcommand{\thefigure}{E.16}
    \centering
    \includegraphics[width=0.7\linewidth]{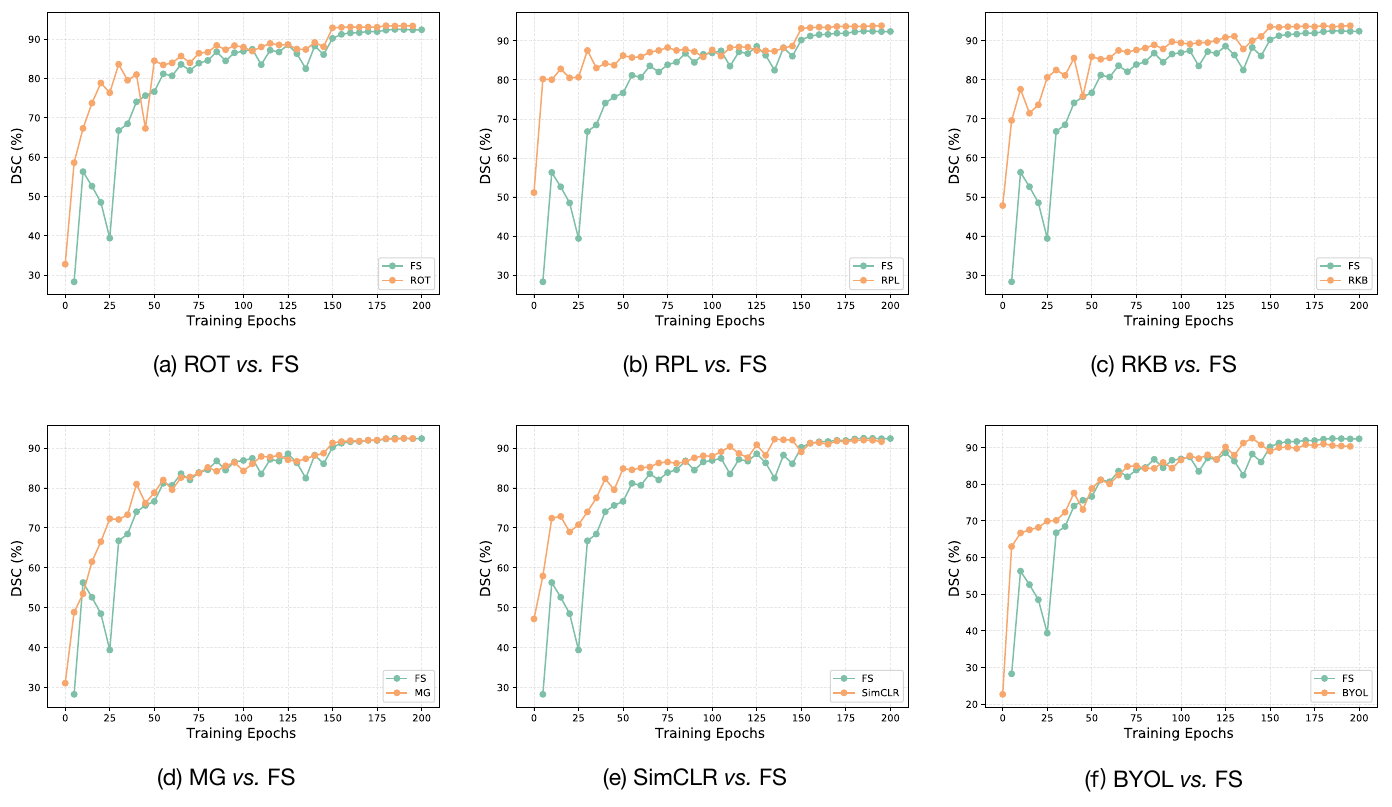}
    \caption{Comparisons of convergence speed between self-supervised pretraining methods (orange) and the from-scratch baseline (green) in LCS.}
    \label{fig:lcs_convergency}
\end{figure*}
\begin{figure*}[!ht]
\renewcommand{\thefigure}{E.17}
    \centering
\includegraphics[width=0.7\linewidth]{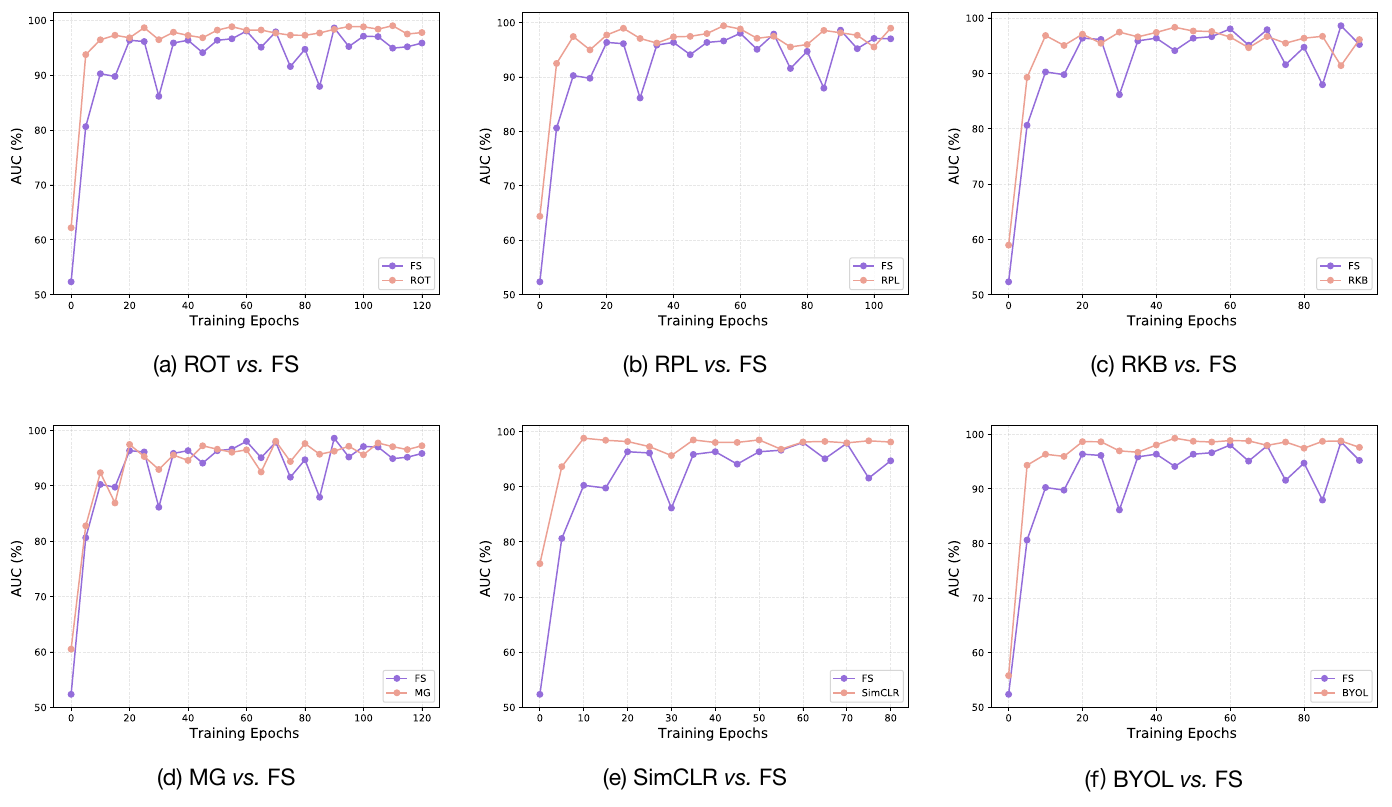}
    \caption{Comparisons of convergence speed between self-supervised pretraining methods (pink) and the from-scratch baseline (purple) in NCC.}
    \label{fig:ncc_convergency}
\end{figure*}
\subsection{The architectures of Jigsaw, RKB and RKB+}
In the process of implementing Jigsaw-puzzling-based methods, we find that the depth of network heads needs to be chosen carefully otherwise inappropriate architecture could cause model collapse. Taking RKB as an example, we report the head architectures and the corresponding validation accuracy of auxiliary sub-tasks in Table~\ref{tab:head_arch}. Even a little difference in the head architectures yields completely different outcomes. An obvious consequence is that the model with deeper classifier layers does not converge and achieves a high error rate on the validation set. Such a phenomenon implies that the manually designed SSL multi-task might lay many pitfalls for user training.
\begin{table}[!h]
 \renewcommand{\thetable}{E.11} 
\renewcommand{\arraystretch}{0.7}
\caption{The comparison of using the online encoder and target encoder in BYOL as the pretrained encoder across three target tasks.}
\centering
\resizebox{0.48\textwidth}{!}{
\begin{tabular}{cccc}
\hline
Encoder & NCC (AUC / \%) & NCS (DSC / \%) & LCS (DSC / \%)  \bigstrut\\ \hline
Online    & 99.52          & 76.13          & 93.33          \bigstrut\\ \hline
Target & 98.64          & 75.37          & 94.43           \bigstrut\\ \hline
\end{tabular}}
\label{tab:BYOL}
\end{table}
\subsection{The online and target encoder in BYOL.} 
There are two encoders in BYOL, where the online encoder is updated by backward propagation and the target encoder is updated through exponential momentum average. It is a common consensus that the features in the target encoder are smoother and preferable when transferred to the downstream tasks. However, our results in Table~\ref{tab:BYOL} suggest that the online encoder outperforms the target encoder in NCC and NCS.
\subsection{The t-SNE visualizations of representations from different models in NCC.}
\label{appen:t_sne}
To better understand how SSL methods improve the performance of NCC over the baseline model, we leverage t-SNE \citep{van2008visualizing} to visualize the features before the last classifier layer. The t-SNE visualizations in Fig.~\ref{fig:tsne_ncc_appen} qualitatively indicate that RPL, RKB+ and BYOL improve the class separability in the embedding space, accounting for their gains in the performance of NCC.
\subsection{Detailed convergence speed results}
From Fig.~\ref{fig:ncc_convergency} and Fig.~\ref{fig:lcs_convergency}, we find that most SSL methods facilitate more steady training and faster convergence speed, especially RPL and SimCLR. At the beginning epochs, most SSL methods surpass the baseline, manifesting the benefits of pretrained representations in downstream tasks.

\section{Proof of the classification theoretical model}

\subsection{Proof of Theorem 1}
\label{sec:proof_theorem1}
Given two Gaussian distributions $Z^+ \sim N(\mu_1, \sigma^2)$ and $Z^- \sim N(\mu_2, \sigma^2)$, the sum of independent Gaussian variables follows the below Gaussian distribution:
\begin{equation}       
	\frac{\sum_{k=1}^{N_+}Z_k^+}{N^+}+\frac{\sum_{k=1}^{N^-}Z_k^-}{N^-} \sim N(\mu_1 + \mu_2, (\frac{1}{N^+}+\frac{1}{N^-})\sigma^2)       
\end{equation}
As introduced in \cite{boucheron2013concentration}, the Gaussian concentration inequality is:
\begin{equation}\label{eq:gaussian_concentration}   
	\mathbb{P}(|Z-E(Z)|\geq \delta) \leq 2e^{\frac{-\delta^2}{2\sigma^2}}
\end{equation}      
where $Z-E(Z)$ is said to be a mean-zero Gaussian variable with variance $\sigma^2$ 
and $\delta > 0$. 
By applying Eq.~\ref{eq:gaussian_concentration} to the term $\frac{\sum_{k=1}^{N^+}Z_k^+}{N^+}+\frac{\sum_{k=1}^{N^-}Z_k^-}{N^-}$, we have:
\begin{equation}       
	\mathbb{P}(|\frac{\sum_{k=1}^{N^+}Z_k^+}{N^+}+\frac{\sum_{k=1}^{N_-}Z_k^-}{N^-}-(\mu_1 + \mu_2)| \geq \delta) \leq 2e^{\frac{-\delta^2}{2\sigma^2}\frac{1}{\frac{1}{N^+}+\frac{1}{N^-}}}
\end{equation}
Therefore, the estimated $\hat\theta_1 = \frac{\sum_{k=1}^{N^+}Z_k^+/N^++\sum_{k=1}^{N^-}Z_k^-/N^-}{2}$ obeys:
\begin{equation}       
	\mathbb{P}(|\hat\theta_1-\frac{\mu_1 + \mu_2}{2}| \geq \frac{\delta}{2}) \leq 2e^{\frac{-\delta^2}{2\sigma^2}\frac{1}{\frac{1}{N^+}+\frac{1}{N^-}}}
\end{equation}
Let $t =\frac{\delta}{2}$, we finally obtain the Theorem 1.
\subsection{Proof of Theorem 2}
\label{sec:proof_theorem2}
Consider a binary classification problem of $\omega_1$ and $\omega_2$, we set $t=\ln{\frac{p(\omega_1)}{p(\omega_2)}}$ and then the Bayesian decision function is:
\begin{equation} 
	\label{eq:bayes}      
	\left\{              
	  \begin{array}{l}   
		h(x) = -\ln{\frac{p(x|w_1)}{p(x|w_2)}} < t, x \in \omega_1 \\ 
		h(x) = -\ln{\frac{p(x|w_1)}{p(x|w_2)}} > t, x \in \omega_2 \\
		
	  \end{array}
	\right.                 
	\end{equation}
in which $h$ is the so-called minus-log-likelihood ratio. Then, we introduce the moment generating function of $h$:
\begin{equation}\label{eq:phi}       
	\phi_1(s) = \int_{-\infty}^{\infty}e^{sh}p(h|\omega_1)dh, ~~s \in [0, 1]
\end{equation}
where we can regard $p(h|\omega_1)$ as the density function of $h$ for $x$ from $\omega_1$. We note that $u(s) = -\ln{\phi_1(s)}$ and define another function:
\begin{equation}\label{eq:pg}     
	p_g(g=h|\omega_1) = \frac{e^{sh}p(h|\omega_1)}{\phi_1(s)}
\end{equation}
Obviously, $\int_{-\infty}^{\infty}p_g(g=h|\omega_1)dh=1$. Rearranging Eq.~\ref{eq:pg} and replacing $\phi_1(s)$ with $u(s)$ gives:
\begin{equation}       
	p(h|\omega_1) = e^{-u(s)-sh}p_g(g=h|\omega_1)
\end{equation}
From Eq.~\ref{eq:bayes}, we can see that x is classified as $\omega_2$ when $h \in (t, \infty)$. The error rate $\epsilon_1$ for $\omega_1$ is defined as the probability of classifying $x$ actually from $\omega_1$ into $\omega_2$:
\begin{equation} \label{eq:error_rate1}    
	\epsilon_1 = \int_{t}^{\infty}p(h|\omega_1)dh=e^{-u(s)}\int_{t}^{\infty}e^{-sh}p_g(g=h|\omega_1)dh
\end{equation}
For $h \in (t, \infty)$ and $s \in [0, 1]$,  $e^{-sh} \leq e^{-st}$. We can obtain the upper bound on $\epsilon_1$ by applying this inequality to Eq.~\ref{eq:error_rate1}:
\begin{equation} \label{eq:epsilon1}          
	\epsilon_1 \leq e^{-u(s)-st}
\end{equation}
Similarly, we can derive the upper bound on $\epsilon_2$ as:
\begin{equation} \label{eq:epsilon2}      
	\epsilon_2 \leq e^{-u(s)+(1-s)t}
\end{equation}
Unlike Eq.~\ref{eq:phi}, we re-express $\phi_1(s)$ using $x$ as the variable:
\begin{equation} \label{eq:phi_v2}      
\phi_1(s) = \int_{\Gamma}e^{sh}p(x|\omega_1)dx = \int_{\Gamma}p(x|\omega_1)^{1-s}p(x|\omega_2)^sdx
\end{equation}
in which, $\Gamma$ is the distribution space of $x$. From Eq.~\ref{eq:phi_v2}, we obtain a new expression of $u(s)$:
\begin{equation} \label{eq:us}      
	u(s) = -\ln\phi_1(s)= -\ln\int_{\Gamma}p(x|\omega_1)^{1-s}p(x|\omega_2)^sdx
	\end{equation}
Since the optimal value of $s$ is complicated to acquire, we  set $s=1/2$. As the result, Eq.~\ref{eq:us} turns to the Bhattacharyya distance between $\omega_1$ and $\omega_2$, i.e. $u(1/2)=D_B(\omega_1, \omega_2)$. For a continuous distribution, the definition of $D_B(\omega_1, \omega_2)$ is:
\begin{equation} \label{eq:DB}      
	D_B(\omega_1, \omega_2) = -\ln\int_{\Gamma}\sqrt{p(x|\omega_1)p(x|\omega_2)}dx
	\end{equation}
This means that we can use the Bhattacharyya distance to compute the error upper bound of the Bayesian classifier. If $x|\omega_1 \sim N(\mu_1, \sigma^2)$ and $x|\omega_2 \sim N(\mu_2, \sigma^2)$, $D_B(\omega_1, \omega_2)=\frac{1}{8}\frac{(\mu_2-\mu_1)^2}{\sigma^2}$ according to \cite{fukunaga2013introduction}.
Let $\lambda={\frac{p(\omega_1)}{p(\omega_2)}}$, $t=\ln\lambda$. The Eq.~\ref{eq:epsilon1} and Eq.~\ref{eq:epsilon2} can be expressed as:
\begin{equation} 
	\label{eq:upper_bound}      
	\left\{              
	  \begin{array}{l}   
			\epsilon_1 \leq e^{-u(1/2)-(1/2)t} = \lambda^{-1/2}e^{-D_B} \\ 
		\epsilon_2 \leq e^{-u(1/2)+(1/2)t} = \lambda^{1/2}e^{-D_B} \\	
	  \end{array}
	\right.                 
	\end{equation}
Finally, we prove Theorem 2 above.

\bibliographystyle{model2-names.bst}
\biboptions{authoryear}
\bibliography{references}

\end{document}